\documentclass{article}

\usepackage[final, nonatbib]{neurips_2021}

\usepackage[utf8]{inputenc} 
\usepackage[T1]{fontenc}    
\usepackage{hyperref}       
\usepackage{url}            
\usepackage{booktabs}       
\usepackage{amsfonts}       
\usepackage{nicefrac}       
\usepackage{microtype}      
\usepackage{xcolor}         
\usepackage{graphicx}
\usepackage{caption}
\usepackage{subcaption}
\usepackage{todonotes}
\usepackage{amsmath}
\usepackage[ruled,noend]{algorithm2e}
\usepackage{wrapfig}
\usepackage[numbers]{natbib}
\usepackage{multibib}
\newcites{a}{References} 

\newcommand{\pluseq}{\mathrel{+}=}

\newcommand{\expnumber}[2]{{#1}\mathrm{e}{#2}}

\title{Collaborating with Humans without Human Data}

\author{
  DJ Strouse\thanks{Equal contribution.}, Kevin R. McKee, Matt Botvinick, Edward Hughes, Richard Everett\footnotemark[1] \\
  DeepMind \\
  \texttt{\{strouse, kevinrmckee, botvinick, edwardhughes, reverett\}@deepmind.com}
}

\begin{document}

\maketitle

\begin{abstract}
  Collaborating with humans requires rapidly adapting to their individual strengths, weaknesses, and preferences. Unfortunately, most standard multi-agent reinforcement learning techniques, such as self-play (SP) or population play (PP), produce agents that overfit to their training partners and do not generalize well to humans. Alternatively, researchers can collect human data, train a human model using behavioral cloning, and then use that model to train ``human-aware'' agents (``behavioral cloning play'', or BCP). While such an approach can improve the generalization of agents to new human co-players, it involves the onerous and expensive step of collecting large amounts of human data first. Here, we study the problem of how to train agents that collaborate well with human partners without using human data. We argue that the crux of the problem is to produce a diverse set of training partners. Drawing inspiration from successful multi-agent approaches in competitive domains, we find that a surprisingly simple approach is highly effective. We train our agent partner as the best response to a population of self-play agents and their past checkpoints taken throughout training, a method we call Fictitious Co-Play (FCP). Our experiments focus on a two-player collaborative cooking simulator that has recently been proposed as a challenge problem for coordination with humans. We find that FCP agents score significantly higher than SP, PP, and BCP when paired with novel agent and human partners. Furthermore, humans also report a strong subjective preference to partnering with FCP agents over all baselines.
\end{abstract}

\section{Introduction}
\label{sec:introduction}

Generating agents which collaborate with novel partners is a longstanding challenge for Artificial Intelligence (AI) \citep{bard2020hanabi, dafoe2020cooperative, klien2004ten, mutlu2013coordination}. Achieving ad-hoc, zero-shot coordination \citep{hu2020otherplay, stone2010adhoc} is especially important in situations where an AI must generalize to novel human partners \citep{bauer2008human, schurr2005coordination}. Many successful approaches have employed human models, either constructed explicitly \citep{choudhury2019utility, javdani2015shared, nikolaidis2013team} or learnt implicitly \citep{carroll2019overcooked, sadigh2018cars}. By contrast, recent work in competitive domains has shown that it is possible to reach human-level using model-free reinforcement learning (RL) without human data, via self-play \citep{brown2018libratus, brown2019pluribus, silver2017alphagozero, silver2018alphazero}. This begs the question: Can model-free RL without human data generate agents that can collaborate with novel humans?

We seek an answer to this question in the space of common-payoff games, where all agents work towards a shared goal and receive the same reward. Self-play (SP), in which an agent learns from repeated games played against copies of itself, does not produce agents that generalize well to novel co-players \cite{bullard2020communication, bullard2021equivalence, foerster2019bad, lowe2020emergent}. Intuitively, this is because agents trained in self-play only ever need to coordinate with themselves, and so make for brittle and stubborn collaborators with new partners who act differently. Population play (PP) trains a population of agents, all of whom interact with each other \citep{lanctot2017unified}. While PP can generate agents capable of cooperation with humans in competitive team games \citep{jaderberg2019human}, it still fails to produce robust partners for novel humans in pure common-payoff settings \citep{carroll2019overcooked}. PP in common-payoff settings naturally encourages agents to play the same way, reducing strategic diversity and producing agents not so different from self-play \citep{garnelo2021diversity}.

Our approach starts with the intuition that the key to producing robust agent collaborators is exposure to diverse training partners. We find that a surprisingly simple strategy is effective in generating sufficient diversity. We train $N$ self-play agents varying only their random seed for neural network initialization. Periodically during training, we save agent ``checkpoints'' representing their strategy at that point in time. Then, we train an agent partner as the best-response to both the fully-trained agents and their past checkpoints. The different checkpoints simulate different skill levels, and the different random seeds simulate breaking symmetries in different ways. We refer to this agent training procedure as \textbf{Fictitious Co-Play (FCP)} for its relationship to fictitious self-play \citep{brown1951fictitious, heinrich2016nfsp, heinrich2015fsp, vinyals2019alphastar}.

We evaluate FCP in a fully-observable two-player common-payoff collaborative cooking simulator. Based on the game Overcooked \citep{overcooked}, it has recently been proposed as a coordination challenge for AI \citep{carroll2019overcooked, mckee2021quantifying, wang2020overcooked}. State-of-the-art performance in producing agents capable of generalization to novel humans was achieved in \citep{carroll2019overcooked} via behavioral cloning (BC) of human data. More precisely, BC was used to produce models that can stand in as human proxies during training in simulation, a method we call behavioral cloning play (BCP). We demonstrate that FCP outperforms BCP in generalizing to both novel agent and human partners, and that humans express a significant preference for partnering with FCP over BCP. Our method avoids the cost and potential privacy concerns of collecting human data for training, while achieving better outcomes for humans at test time.

\begin{figure}[t]
     \centering
     \includegraphics[width=0.95\textwidth]{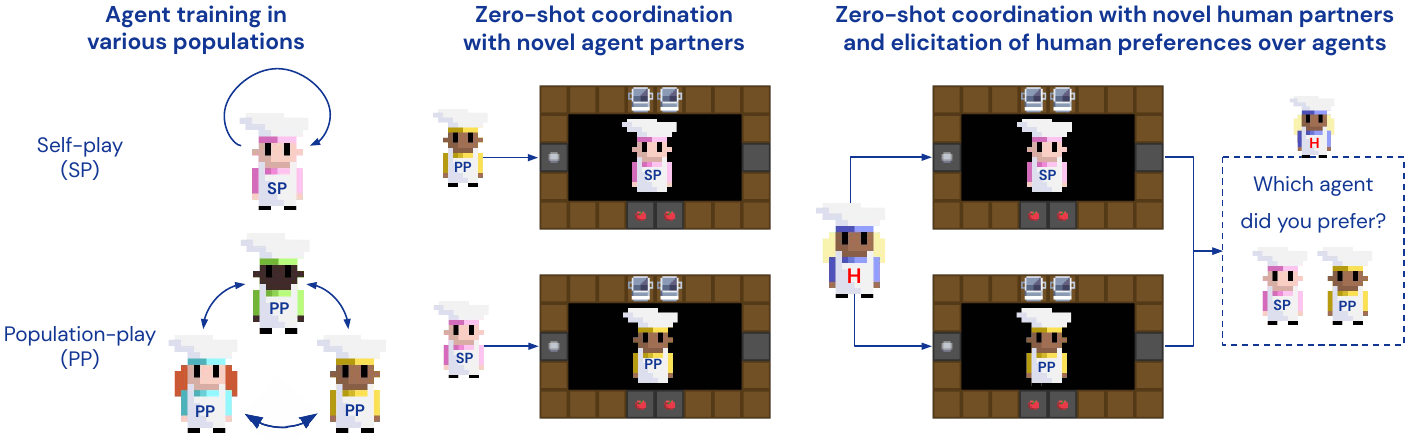}
     \caption{In this work, we evaluate a variety of agent training methods (Section~\ref{sec:methods}) in zero-shot coordination with agents (Section~\ref{sec:agent_agent}). We then run a human-agent collaborative study designed to elicit human preferences over agents (Section~\ref{sec:human_agent}).}
     \label{fig:paper_overview}
     \vspace{-1em}
\end{figure}

We summarize the novel contributions of this paper as follows:

\begin{enumerate}
  \itemsep0.1em 
  \item We propose Fictitious Co-Play (FCP) to train agents capable of zero-shot coordination with humans (Section \ref{sec:methods/fcp}).
  \item We demonstrate that FCP agents generalize better than SP, PP, and BCP in zero-shot coordination with a variety of held-out agents (Section \ref{sec:agent_agent/results}).
  \item We propose a rigorous human-agent interaction study with behavioral analysis and participant feedback (Section \ref{sec:human_agent/methods}).
  \item We demonstrate that FCP significantly outperforms the BCP state-of-the-art, both in task score and in human partner preference (Section \ref{sec:human_agent/results}). 
\end{enumerate}

\section{Methods}
\label{sec:methods}

\begin{figure}[h]
    \centering
    \includegraphics[width=0.95\linewidth]{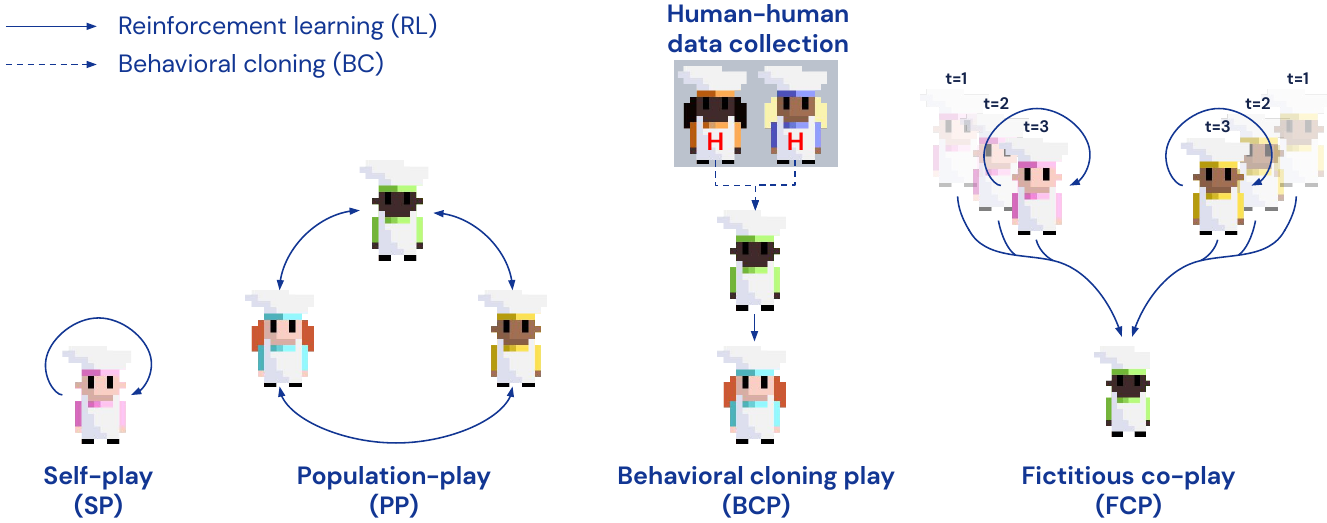}
    \caption{The four agent training methods we evaluate in this work. \textbf{Self-play (SP)} where an agent learns with itself, \textbf{population-play (PP)} where a population of agents are co-trained together, and \textbf{behavioral cloning play (BCP)} where data from human games is used to create a behaviorally cloned agent with which an RL agent is then trained. In our method, \textbf{Fictitious Co-Play (FCP)}, $N$ self-play agents are trained independently and checkpointed throughout training. An agent is then trained to best respond to the entire population of SP agents and their checkpoints.}
    \label{fig:algorithms}
    \vspace{-1em}
\end{figure}

\subsection{Fictitious Co-Play (FCP)}
\label{sec:methods/fcp}

Diverse training conditions have been shown to make agents more robust, from environmental variations (i.e. domain randomization \citep{openai2019rubiks, peng2018domain, tobin2017domain}) to heterogeneity in training partners \citep{vinyals2019alphastar}. We seek to train agents that are robust partners for humans in common-payoff games, and so extend this line of work to that setting.

One important challenge in collaborating with novel partners is dealing with symmetries \citep{hu2020otherplay}. For example, two agents A and B facing each other may move past each other by A going left and B going right, or vice versa. Both are valid solutions, but a good agent partner will adaptively switch between these conventions if a human clearly prefers one over the other. A second important challenge is dealing with variations in skill level. Good agent partners should be able to assist both highly-skilled partners, as well as partners who are still learning.

Fictitious co-play (FCP) is a simple two-stage approach for training agents that overcomes both of these challenges (Figure~\ref{fig:algorithms}, right). In the first stage, we train a diverse pool of partners. To allow the pool to represent different symmetry breaking conventions, we train $N$ partner agents in self-play. Since these partners are trained independently, they can arrive at different arbitrary conventions for breaking symmetries. To allow the pool to represent different skill levels, we use multiple checkpoints of each self-play partner throughout training. The final checkpoint represents a fully-trained ``skillful'' partner, while earlier checkpoints represent less skilled partners. Notably, by using multiple checkpoints per partner, this additional diversity in skill incurs no extra training cost.

In the second stage, we train an FCP agent as the best response to the pool of diverse partners created in the first stage. Importantly, the partner parameters are frozen and thus FCP must learn to adapt to partners, rather than expect partners to adapt to it. In this way, FCP agents are prepared to follow the lead of human partners, and learn a general policy across a range of strategies and skills. We call our method ``fictitious'' co-play for its relationship to fictitious self-play in which competitive agents are trained with past checkpoints (in that case, to avoid strategy cycling) \citep{brown1951fictitious, heinrich2016nfsp, heinrich2015fsp, lanctot2017unified, vinyals2019alphastar}.

\subsection{Baselines and ablations}
\label{sec:methods/baselines_and_ablations}
We compare FCP agents to the three baseline training methods listed below, each varying only in their set of training partners, with the RL algorithm and architecture consistent across all agents:

\begin{enumerate}
    \item Self-play (SP), where agents learn solely through interaction with themselves.
    \item Population-play (PP), where a population of agents are co-trained through random pairings.
    \item Behavioral cloning play (BCP), where an agent is trained with a BC model of a human \citep{carroll2019overcooked}.
\end{enumerate}

We also evaluate three variations on FCP to better understand the conditions for its success:

\begin{enumerate}
    \item To test the importance of including past checkpoints in training, we evaluate an ablation of FCP in which agents are trained only with the converged checkpoints of their partners (FCP$_{-T}$ for ``FCP minus time'').
    \item To test whether FCP would benefit from additional diversity in its partner population, we evaluate an augmentation of FCP in which the population of SP partners varies not just in random seed, but also in architecture (FCP$_{+A}$ for ``FCP plus architectural variation'').
    \item To test whether architectural variation can serve as a full replacement for playing with past checkpoints, we evaluate the combination of both modifications (FCP$_{-T,+A}$).
\end{enumerate}

\subsection{Environment}
\label{sec:methods/environment}

\begin{figure}[h]
     \centering
     \includegraphics[width=0.94\textwidth]{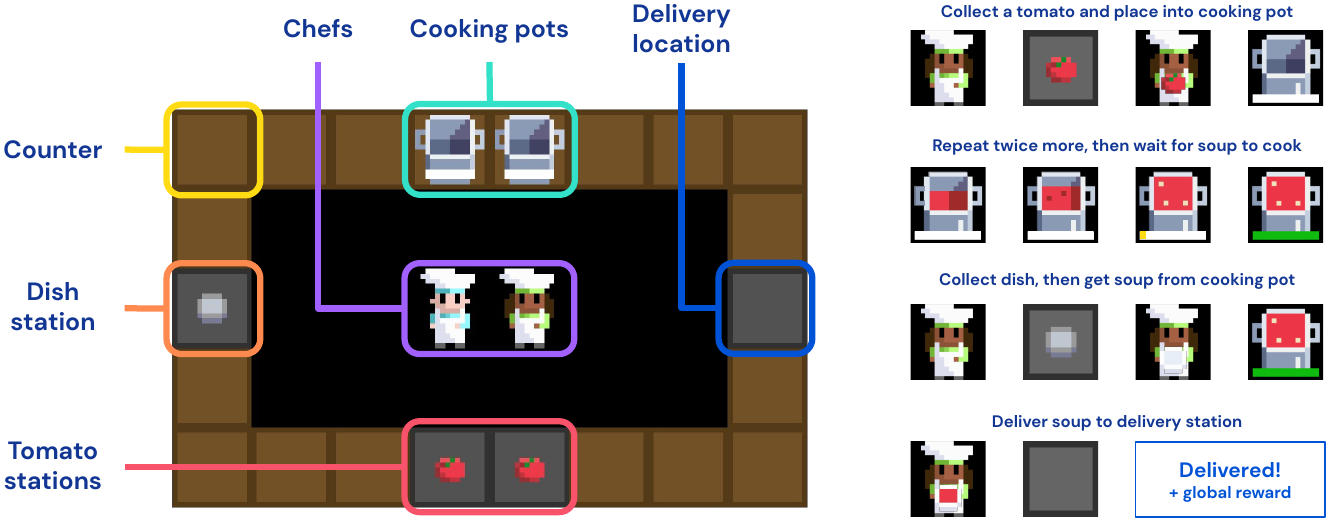}
     \caption{\textbf{The Overcooked environment:} a two-player common-payoff game in which players must coordinate to cook and deliver soup.}
     \label{fig:environment}
     \vspace{-1em}
\end{figure}

Following prior work on zero-shot coordination in human-agent interaction, we study the Overcooked environment (see Figure~\ref{fig:environment}) \citep{carroll2019overcooked, charakorn2020diversification, knott2021tests, mckee2021quantifying, wang2020overcooked}. We draw particular inspiration from the environment in \citet{carroll2019overcooked}. For full details, see Appendix~\ref{sec:app/environment}.

In this environment, players are placed into a gridworld kitchen as chefs and tasked with delivering as many cooked dishes of tomato soup as possible within an episode. This involves a series of sequential high-level actions to which both players can contribute: collecting tomatoes, depositing them into cooking pots, letting the tomatoes cook into soup, collecting a dish, getting the soup, and delivering it. Upon a successful delivery, both players are rewarded equally.

To effectively complete the task, players must learn to navigate the kitchen and interact with objects in the correct order, all while maintaining awareness of their partner's behavior to coordinate with them. This environment therefore presents the challenges of both movement and strategic coordination.

Each player observes an egocentric RGB view of the world, and at every step can perform one of six actions: \texttt{stand still},\texttt{ move \{up, down, left, right\}},\texttt{ interact}. The behavior of \texttt{interact} varies based on the cell which the player is facing (e.g. place tomato on counter).

\begin{figure}[h]
    \centering
     \includegraphics[width=0.94\textwidth]{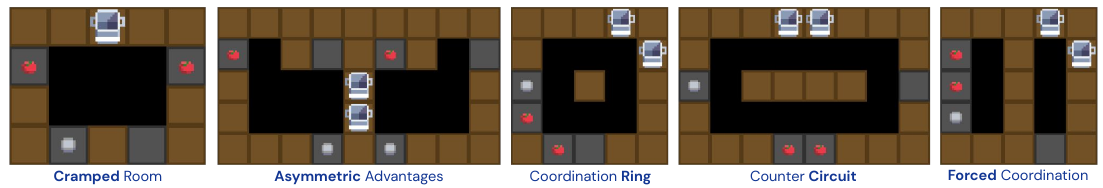}
     \caption{\textbf{Layouts:} the kitchens which agents and humans play in, each emphasizing different coordination strategies. Highlighted in bold are the terms used to refer to each in the rest of this paper.}
     \label{fig:environment/layouts}
\end{figure}

\subsection{Implementation details}
\label{sec:methods/implementation}

Here we highlight several key implementation details for our training methods. For full details, including the architectures, hyperparameters, and compute used, please see Appendix~\ref{sec:app/agent}.

For our reinforcement learning agents, we use the V-MPO \citep{song2019vmpo} algorithm along with a ResNet \citep{he2016resnet} plus LSTM \citep{hochreiter1997lstm} architecture which we found led to optimal behavior across all layouts. Agents are trained using a distributed set of environments running in parallel \citep{espeholt2018impala}, each sampling two agents from the training population to play together every episode.

Both PP and FCP are trained with a population size of $N=32$ agents which are sampled uniformly. For FCP, we use 3 checkpoints for each agent, therefore incurring no additional training burden: (1) at initialization (i.e. a low-skilled agent), (2) at the end of training (i.e. a fully-trained expert agent), and (3) at the middle of training, defined as when the agent reaches 50\% of its final reward (i.e. an average-skilled agent). When varying architecture for the training partners of the FCP$_{+A}$ and FCP$_{-T,+A}$ variants, we vary whether the partners use memory (i.e. LSTM vs not) and the width of their policy and value networks (i.e. 16 vs 256). In total, we train 8 agents for each of the 4 combinations, leaving the total population size of $N=32$ unchanged, ensuring a fair comparison.

To train agents via behavioral cloning \citep{pomerleau1991efficient}, we use the open-source Acme \citep{hoffman2020acme} to learn a policy from human gameplay data. Specifically, we collected 5 human-human trajectories of length 1200 time steps for each of the 5 layouts, resulting in 60k total environment steps. We divide this data in half and train two BC agents: (1) a partner for training a BCP agent, and (2) a ``human proxy'' partner for agent-agent evaluation. Following \citet{carroll2019overcooked}, we use a set of feature-based observations for the agents (as opposed to RGB) and generate comparable results: performance is higher on 3 layouts (\texttt{asymmetric}, \texttt{cramped}, and \texttt{ring}) but poorer on the other 2 (\texttt{circuit} and \texttt{forced}).

\section{Related work}
\label{sec:related_work}

\textbf{Ad-hoc team play}\quad There is a large and diverse body of literature on ad-hoc team-play \citep{barrett2017plastic, stone2010adhoc}, also known as zero-shot coordination \citep{hu2020otherplay}. Prior work based in game-theoretic settings has suggested the benefits of planning \cite{wu2011adhoc}, online learning \cite{melo2015adhoc}, and novel solution concepts \cite{albrecht2013adhoc}, to name a few examples. More recently, multi-agent deep reinforcement learning has provided the tools to scale to more complex gridworld or continuous control settings, leading to work on hierarchical social planning \cite{kleiman2016coordinate}, adapting to existing social conventions \cite{lerer2019conventions, shih2021conventions}, trajectory diversity \cite{lupu2021trajedi}, and theory of mind \cite{choudhury2019utility}. Ad-hoc team-play among novel agent partners is also an object of active study in the emergent communication literature \cite{bullard2020communication, bullard2021equivalence, lowe2019learning}. This prior work has tended to focus on generalization to held-out agent partners as a proxy for human co-players. 
 
Collaborative play with novel humans has been evaluated more actively in the context of training agent assistants; see for instance \citep{pilarski2019vr, tylkin2021atari}. To our knowledge, our FCP agents represent the state-of-the-art in coordinating with novel human partners on an equal footing of capabilities in a rich gridworld environment, as measured by the challenge tasks in \citet{carroll2019overcooked}.

\textbf{Diversity in multi-agent reinforcement learning}\quad In multi-agent reinforcement learning, agents that train with behaviorally diverse populations of game partners tend to demonstrate stronger performance than their self-play counterparts. For example, across a range of multi-agent games, generalization to held-out populations can be improved by training larger and more diverse populations \cite{charakorn2020diversification, lowe2017multi, mckee2021quantifying}. In mixed-motive settings, cooperation among agents can be encouraged through social diversity, such as in player preferences and rewards \cite{baker2020rusp, mckee2020social, mckee2021deep}. Similarly, competitiveness can be optimized through selective matchmaking between increasingly diverse agents \cite{garnelo2021diversity, lanctot2017unified, vinyals2019alphastar}.

Despite the increased focus on improving multi-agent performance, evaluation has typically been constrained to agent-agent settings. High-performing agents have infrequently been evaluated with humans, particularly in non-competitive domains \cite{dafoe2020cooperative}. We add to this growing literature, showing that training with diversity is a powerful approach for effective human-agent collaboration.

\textbf{Human-agent interaction}\quad In recent years, increased attention has been directed toward designing machine learning agents capable of collaborating with humans \citep{lockhart2020human, pilarski2019vr, tylkin2021atari, zheng2020economist} (see also \citep{dafoe2020cooperative} for a broader review on Cooperative AI). \citet{tylkin2021atari} is particularly notable in also demonstrating that partially trained agents can be useful learning targets for human helpers, although in a different domain (cooperative Atari). Our method, FCP, can be seen as extending theirs by training with multiple ``skill levels'' and random seeds, rather than just one, which we demonstrate to be crucial to our agents' performance (Tables~\ref{tab:agent_agent/results/ablation} and \ref{tab:agent_agent/results/population_size} and Figure~\ref{fig:human_eval_b}).

A key preceding entry in this research area is \citet{carroll2019overcooked}, who similarly investigated human-agent coordination in Overcooked. We use their method (BCP) as a baseline throughout our experiments (Section~\ref{sec:methods/baselines_and_ablations}). Relative to BCP, our approach removes the need for the expensive step of human data collection for agent training. Furthermore, through our novel human-agent experimental design, we go beyond objective performance metrics to compare the subjective preferences that agents generate. For a detailed comparison of methods and results, see Appendix~\ref{sec:app/related_work}.

\section{Zero-shot coordination with agents}
\label{sec:agent_agent}
In this section, we evaluate our FCP agent, its ablations, and the baselines with held-out agents. 

\subsection{Evaluation method: collaborative evaluation with agent partners}
\label{sec:agent_agent/method}
Our primary concern in this work is generalization to novel \emph{human} partners (as investigated in Section~\ref{sec:human_agent}). However, just as collecting human-human data for behavioral cloning is expensive, so too is evaluating agents with humans. Consequently, we instead use generalization to held-out \emph{agent} partners as a cheap proxy of performance with humans. This is then used to guide our model selection process, allowing us to be more targeted with the agents we select for our human-agent evaluations.

We evaluate with three held-out populations:
\begin{enumerate}
    \item A BC model trained on human data, \textit{H$_{\rm{proxy}}$}, intended as a proxy of generalization to humans, as done by \citet{carroll2019overcooked}.
    \item A set of self-play agents varying in seed, architecture, and training time (specifically, held-out seeds of the $N=32$ partners trained for the FCP$_{+A}$ agent; see Section~\ref{sec:methods/implementation}). These are intended to test generalization to a diverse yet still skillful population.
    \item Randomly initialized agents intended to test generalization to low-skill partners.
\end{enumerate}

For all results, we report the average number of deliveries made by both players within an episode, aggregated across the 5 different layouts from Figure~\ref{fig:environment/layouts} (with the per-layout results reported in Appendix~\ref{sec:app/ai_ai/additional_results}). We estimate mean and standard deviation across 5 random seeds. For each seed, we evaluate the agent with all members of the held-out population for 10 episodes per agent-partner pair.

\subsection{Results}
\label{sec:agent_agent/results}

\subsubsection*{Finding 1: FCP significantly outperforms all baselines} 
\begin{figure}[h]
     \centering
     \begin{subfigure}[t]{0.3\textwidth}
         \centering
         \hspace*{-.5cm}\includegraphics[width=\textwidth]{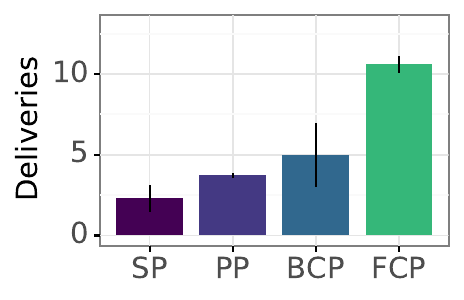}
         \vspace{-.1cm}
         \caption{With \textit{H$_{\rm{proxy}}$}.}
         \label{fig:agent_agent/results/crossplay/BC}
     \end{subfigure}
     \hfill
     \begin{subfigure}[t]{0.3\textwidth}
         \centering
         \hspace*{-.5cm}\includegraphics[width=\textwidth]{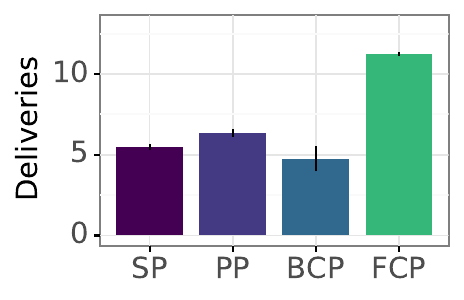}
         \vspace{-.1cm}
         \caption{With diverse SP agents.}
         \label{fig:agent_agent/results/crossplay/RL}
     \end{subfigure}
     \hfill
     \begin{subfigure}[t]{0.3\textwidth}
         \centering
         \hspace*{-.5cm}\includegraphics[width=\textwidth]{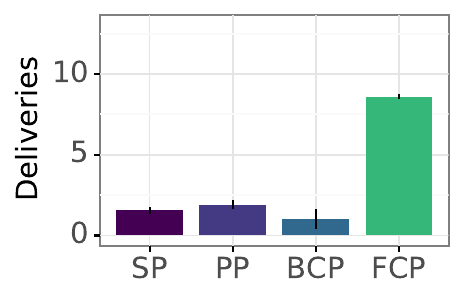}
         \vspace{-.1cm}
         \caption{With random agents.}
         \label{fig:agent_agent/results/crossplay/random}
     \end{subfigure}
    \caption{\textbf{Agent-agent collaborative evaluation:} Performance of each agent when partnered with each of the held-out populations (Section~\ref{sec:agent_agent/method}) in episodes of length $T = 540$. Importantly, FCP scores higher than all baselines with a variety of test partners. Error bars represent standard deviation over five random training seeds. Plots aggregate data across kitchen layouts; results calculated by individual layout can be found in Appendix~\ref{sec:app/ai_ai/additional_results}.}
    \label{fig:agent_agent/results/crossplay}
\end{figure}

\vspace{-0.6em}
To begin, we compare our FCP agent and the baselines when partnered with the three held-out populations introduced above. As can be seen in Figure~\ref{fig:agent_agent/results/crossplay}, FCP significantly outperforms all baselines when partnered with all three held-out populations. Notably, it performs better than BCP with \textit{H$_{\rm{proxy}}$}, even though BCP trains with such a model and FCP does not. Similar to \citet{carroll2019overcooked}, we find that BCP significantly outscores SP.

When paired with a randomly initialized partner which behaves suboptimally, we see an even greater difference between FCP and the baselines. Given that FCP is trained with non-held-out versions of such agents, it may not be surprising that it does so well with partners that behave poorly. However, what is surprising is how brittle the other training methods are. This suggests that they may not perform well with humans who are not highly skilled players, which we will see in Section~\ref{sec:human_agent}.

\subsubsection*{Finding 2: Training with past checkpoints is the most beneficial variation for performance} \vspace{-0.6em}
Next, we investigate how the different training partner variations influence FCP's performance. In particular, we separately ablate the past checkpoints ($T$) and architecture ($A$) variations, evaluating them with the same partners as in Figure~\ref{fig:agent_agent/results/crossplay}. The results of this evaluation are presented in Table~\ref{tab:agent_agent/results/ablation}. 

\begin{table}[h]
    \centering
    \begin{tabular}{ccccc}
        \hline
        Partner & FCP & FCP$_{-T}$ & FCP$_{+A}$ & FCP$_{-T,+A}$ \\ \hline
        \textit{H$_{\rm{proxy}}$} & $10.6\pm0.5$ & $4.7\pm0.4$ & $9.9\pm0.6$ & $7.0\pm0.8$ \\ 
        Diverse SP & $11.2\pm0.1$ & $6.9\pm0.1$ & $11.1\pm0.4$ & $8.6\pm0.4$ \\
        Random & $8.6\pm0.2$ & $1.0\pm0.1$ & $8.4\pm0.4$ & $3.2\pm0.5$ \\         
        \hline
    \end{tabular}
    \vspace{.3cm}
    \caption{\textbf{Ablation results:} Performance of each variation of FCP -- training with past partner checkpoints ($T$ for time) and adding partner variation in architecture ($A$). Scores are mean deliveries with standard deviation over 5 random seeds. Notably, we find that the inclusion of past checkpoints is essential for strong performance (FCP > FCP$_{-T}$), and additionally including architectural variation does not improve performance (FCP $\approx$ FCP$_{+A}$). However, architectural variation is better than no variation, improving performance when past checkpoints are not available (FCP$_{-T,+A}$ > FCP$_{-T}$).}
    \label{tab:agent_agent/results/ablation}
    \vspace{-1em}
\end{table}

Comparing the FCP and FCP$_{-T}$ columns, we see that removing past checkpoints from training significantly reduces performance. Comparing the FCP and FCP$_{+A}$ columns, we see that adding architectural variation to the training population offers no improvement over training with past checkpoints. However, comparing the FCP$_{-T}$ and FCP$_{-T, +A}$ columns, we see that without training with past checkpoints, architectural variation in the population does improve performance.

\section{Zero-shot coordination with humans}
\label{sec:human_agent}

Ultimately, our goal is to develop agents capable of coordinating with novel human partners. In this section, we run an online study to evaluate our FCP agent and the baseline agents in collaborative play with human partners.

\begin{figure}[h]
     \centering
     \begin{subfigure}[b]{\textwidth}
         \centering
         \centerline{\includegraphics[width=1.1\textwidth]{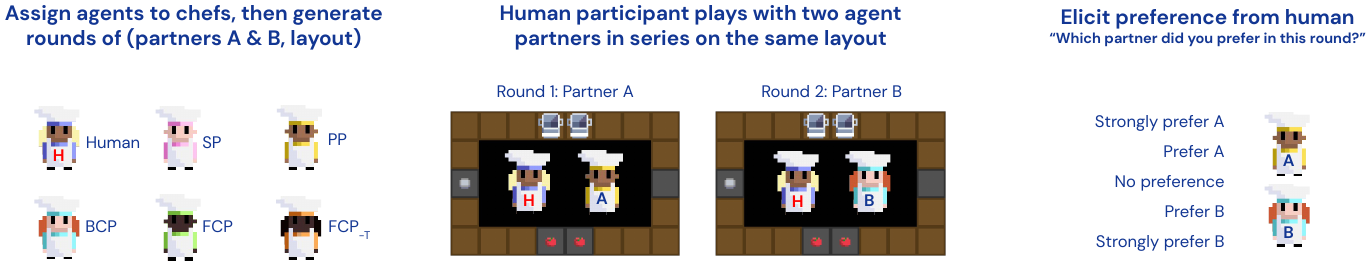}}
     \end{subfigure}
     \vspace{-.3cm}
     \caption{\textbf{Human-agent collaborative study:} For our human-agent collaboration study, we recruited participants online to play games with FCP and baseline agents. Participants played a randomized sequence of episodes with different agent partners and kitchen layouts. After every two episodes, participants reported the direction and strength of their preference between their last two partners.}
     \label{fig:human_agent/methods/method}
\end{figure}

\subsection{Evaluation method: collaborative evaluation with human participants}
\label{sec:human_agent/methods}

To test how effectively FCP's performance generalizes to human partners, we recruited participants from Prolific \citep{eyal2021data, peer2017beyond} for an online collaboration study ($N = 114$; 37.7\% female, 59.6\% male, 1.8\% nonbinary; median age between 25--34 years). We used a within-participant design for the study: each participant played with a full cohort of agents (i.e. generated through every training method). This design allowed us to evaluate both objective performance as well as subjective preferences.

Participants first read game instructions and played a short tutorial episode guiding them through the dish preparation sequence (see Appendix~\ref{sec:app/human_ai/screenshots} for instruction text and study screenshots). Participants then played 20 episodes with a randomized sequence of agent partners and kitchen layouts. Episodes lasted $T = 300$ steps (1 minute) each. After every two episodes, participants reported their preference over the agent partners from those episodes on a five-point Likert-type scale. After playing all 20 episodes, participants completed a debrief questionnaire collecting standard demographic information and open-ended feedback on the study. Our statistical analysis below primarily relies upon the repeated-measures analysis of variance (ANOVA) method. See Appendix~\ref{sec:app/human_ai} for additional details of our study design and analysis, including independent ethical review.

\subsection{Results}
\label{sec:human_agent/results}

\subsubsection*{Finding 1: FCP coordinates best with humans, achieving the highest score across maps} \vspace{-0.6em}

To begin, we compare the objective team performance supported by our FCP and baseline agents. The strong FCP performance observed in agent-agent play generalizes to human-agent collaboration: the FCP-human teams significantly outperform all other agent-human teams, achieving the highest average scores across maps, every $p < 0.001$ (Figure~\ref{fig:human_eval_a}), while performing as well as or better than the other teams on each individual map (see Appendix~\ref{sec:app/human_ai/quantitative_results}). Echoing the results from our agent-agent ablation experiments (Table~\ref{tab:agent_agent/results/ablation}), the inclusion of past checkpoints in training proves critical to FCP's strong performance, $p < 0.001$ (Figure~\ref{fig:human_eval_b}). Similar to \citet{carroll2019overcooked}, we find that BCP outscores SP when collaborating with human players, $p < 0.001$.

\subsubsection*{Finding 2: Participants prefer FCP over all baselines} \vspace{-0.6em}

\begin{figure}[h]
     \centering
     \begin{subfigure}[b]{0.35\textwidth}
         \centering
         \hspace*{-.2cm}\includegraphics[height=9em]{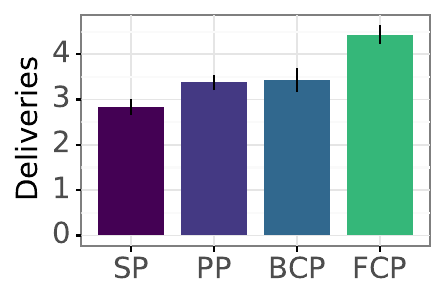}
         \vspace{-.1cm}
         \caption{Number of deliveries by partner (FCP and baselines).}
         \label{fig:human_eval_a}
     \end{subfigure}
     \hfill
     \begin{subfigure}[b]{0.28\textwidth}
         \centering
         \hspace*{-.3cm}\includegraphics[height=9em]{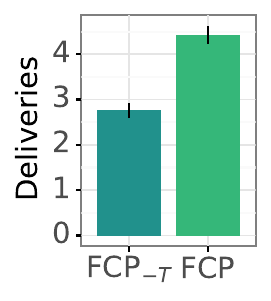}
         \vspace{-.1cm}
         \caption{Number of deliveries by partner (FCP and FCP$_{-T}$).}
         \label{fig:human_eval_b}
     \end{subfigure}
     \hfill
     \begin{subfigure}[b]{0.3\textwidth}
         \centering
         \hspace*{-.3cm}\includegraphics[height=10em]{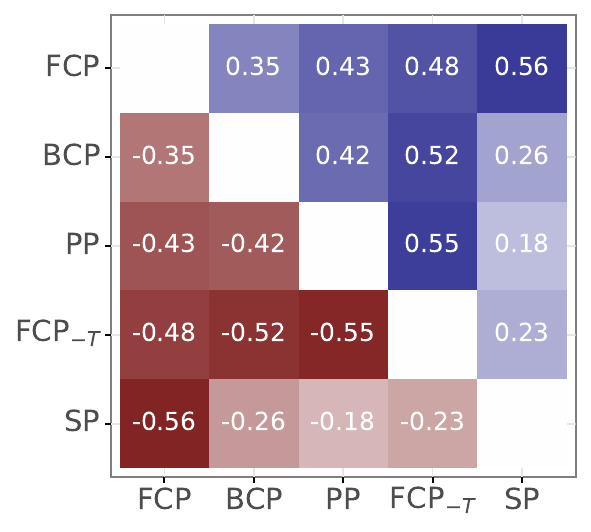}
         \vspace{-.1cm}
         \caption{Participant preference for row partner over column partner.}
        \label{fig:human_eval_c}
     \end{subfigure}
    \caption{\textbf{Human-agent collaborative evaluation}: Evaluation and preference metrics from human-agent play in episodes of length $T = 300$. Error bars represents 95\% confidence intervals, calculated over episodes. Plots aggregate data across kitchen layouts; results calculated by individual layout can be found in Appendix~\ref{sec:app/human_ai/quantitative_results}.}
    \label{fig:human_eval}
\end{figure}

FCP's strong collaborative performance carries over to our participants' subjective partner preferences. Participants expressed a significant preference for FCP partners over all other agents, including BCP, with every $p < 0.05$ (Figure~\ref{fig:human_eval_c}). Notably, while human-BCP and human-PP teams did not significantly differ in their completed deliveries, participants reported significantly preferring BCP over PP, $p = 0.003$, highlighting the informativeness of our subjective analysis.

\subsection{Exploratory behavioral analysis}
\label{sec:human_agent/behavioral_analysis}

To better understand how the human-agent scores and preferences may have arisen, here we analyze the resulting action trajectories of each human and agent player in our experiment.

\begin{figure}[h]
     \centering
     \begin{subfigure}[b]{0.4\textwidth}
         \centering
         \includegraphics[width=\textwidth]{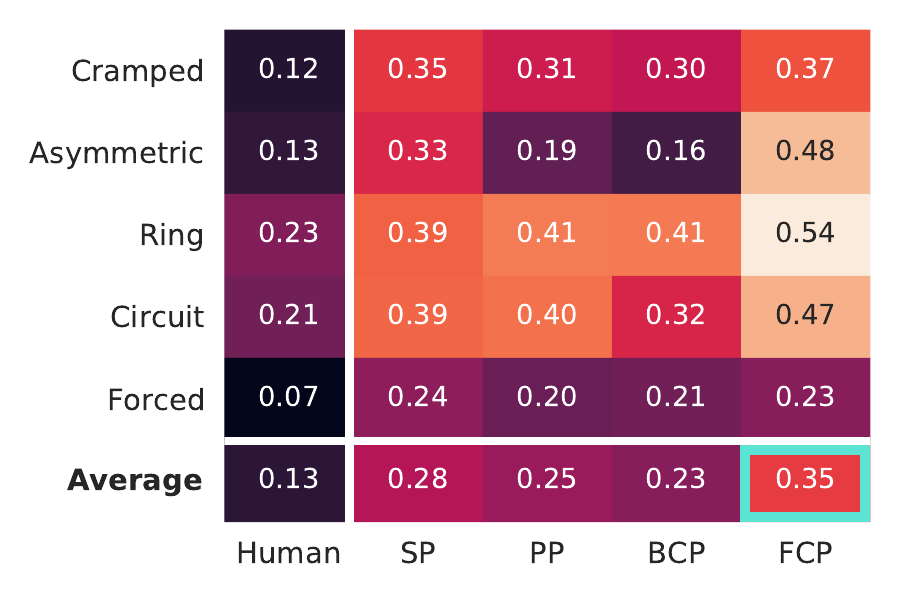}
         \caption{Proportion of episode spent moving.}
         \label{fig:human_ai/behavioral_analysis/time_moving}
     \end{subfigure}
     \hspace{2em}
     \begin{subfigure}[b]{0.4\textwidth}
         \centering
         \includegraphics[width=\textwidth]{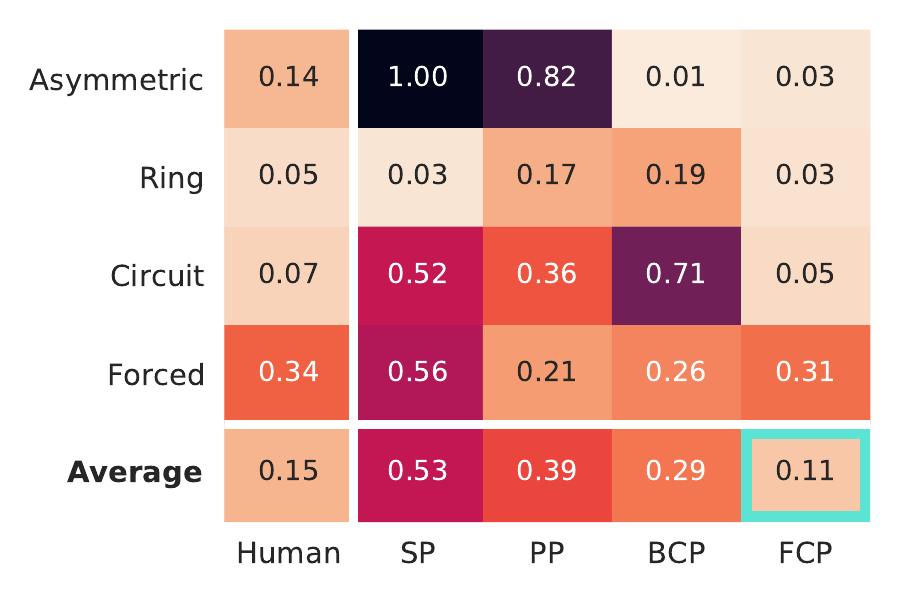}
         \caption{Differences in pot preference.}
         \label{fig:human_ai/behavioral_analysis/pot_preference}
     \end{subfigure}     
     \caption{\textbf{Behavioral analysis:} (a) FCP is able to move most frequently ($35\%$ of the time), corresponding to the best movement coordination with human partners. (b) FCP exhibits the most equal preferences over cooking pots ($0.11$ difference), aligning with human preferences. Values are calculated as the absolute difference in preferences between the two pots; 1 indicates that the player only uses one of the two available pots, while 0 indicates that the player uses both pots equally.}
     \label{fig:human_agent/behavioral_analysis/figure}
     \vspace{-1em}
\end{figure}

\subsubsection*{Finding 1: FCP exhibits the best movement coordination with humans} \vspace{-0.6em}
First, we investigate how much each player moves in an episode (Figure~\ref{fig:human_ai/behavioral_analysis/time_moving}), where moving in a higher fraction of timesteps may suggest fewer collisions and thus better coordination with a partner. Notably, we observe two results: (1) humans rarely move, a behavior which is out-of-distribution for typical training methods (e.g. SP, PP) but is seen in the training distribution for BCP and FCP. (2) FCP moves the most on all layouts other than Forced, suggesting it is better at coordinating its movement strategy with its partner. This result was also reported by human participants, for example:  ``I noticed that some of my partners seemed to know they needed to move around me, while others seemed to get `stuck' until I moved out of their way'' (see Appendix~\ref{sec:app/human_ai} for more examples).

\vspace{-0.1em}  
\subsubsection*{Finding 2: FCP's preferences over cooking pots aligns best with that of humans} \vspace{-0.6em}
Next, we investigate whether there was a preference for a specific cooking pot in the layouts which included two cooking pots (Figure~\ref{fig:human_ai/behavioral_analysis/pot_preference}). To do this, we calculate the difference in the number of times each pot was used by each player, where a high value indicates a strong preference for one pot and a low value indicates more equal preference for the two pots.

As can be seen in the FCP column, our agent typically has the most aligned preferences with that of humans ($0.11$ for FCP to $0.14$ for humans). Behaviorally speaking, this means that our agent prefers one cooking pot over the other $55.5\%$ of the time (i.e. a $0.11$ point difference). In contrast, all other agents have a strong preference for a single pot. This is a non-adaptive strategy which generalizes poorly to typical human behavior of using both pots, leading to worse performance.

\vspace{-0.5em}
\section{Discussion}
\label{sec:discussion}
\vspace{-0.6em}
\textbf{Summary}\quad In this work, we investigated the challenging problem of zero-shot collaboration with humans without using human data in the training pipeline. To accomplish this, we introduced Fictitious Co-Play (FCP) -- a surprisingly simple yet effective method based on creating a diverse set of training partners. We found that FCP agents scored significantly higher than all baselines when partnered with both novel agent and human partners. Furthermore, through a rigorous human-agent experimental design, we also found that humans reported a strong subjective preference to partnering with FCP agents over all baselines.

\textbf{Limitations and future work}\quad Our method currently relies on the manual process of initially training and selecting a diverse set of partners. This is not only time consuming, but also prone to researcher biases that may negatively influence the behavior of the created agents. Additionally, while we found FCP with a partner population size of $N=32$ sufficient here, for more complex games, FCP may require an unrealistically large partner population size to represent sufficiently diverse strategies. To address these concerns, methods for automatically generating partner diversity for common-payoff games may be important. Possibilities include adaptive population matchmaking as been used in competitive zero-sum games \citep{vinyals2019alphastar}, as well as auxiliary objectives that explicitly encourage behavioral diversity \citep{eysenbach2018diayn, lupu2021trajedi, mahajan2019maven}.

Our method requires a known and fixed reward function. We also focus on one domain in order to compare with prior work which has argued that human-in-the-loop training is necessary. Consequently, the resulting agents are only designed to adaptively collaborate on a single task, and not to infer human preferences in general \citep{abramson2020interactive, ibarz2018preferences, russell2019human}. Moreover, if a task's reward function is poorly aligned with how humans approach the task, our method may well produce subpar partners, as would any method without access to human data. Thus, additional domains and tasks should be studied to better understand how our method generalizes. Targeted experiments to test specific forms of generalization may be especially helpful in this regard \citep{knott2021tests}, as could approaches that procedurally generate environment layouts requiring diverse solutions \citep{fontaine2021procedural}.

Finally, it may be possible to produce even stronger agent assistants by combining the strengths of FCP (i.e. diversity) and BCP (i.e. human-like play). Indeed, \citet{knott2021tests} recently demonstrated that modifying BCP to train with \emph{multiple} BC partners produces more robust collaboration with held-out agents, a finding that would be interesting to test with human partners.

\textbf{Societal impact}\quad A challenge for this line of work is ensuring agent behavior is aligned with human values (i.e. the AI value alignment problem \citep{gabriel2020alignment, russell2019human}). Our method has no guarantees that the resulting policy aligns with the preferences, intentions, or welfare of its potential partners. It likewise does not exclude the possibility that the target being optimized for is harmful (e.g. if the agent's partner expresses preferences or intentions to harm others). This could therefore produce negative societal effects either if training leads to poor alignment or if agents are optimized for harmful metrics.

One potential strategy for mitigating these risks is the use of human preference data \citep{christiano2017deep}. Such data could be used to fine-tune and filter trained agents before deployment, encouraging better alignment with human values.
A key question in this line of research is how human preference data should be aggregated---or selected, in the case of expert preferences---when our aim is to create socially aligned agents (i.e. agents that are sufficiently aligned for everyone).
Relatedly, targeted research on human beliefs and perceptions of AI \cite{mckee2020impressions}, and how they steer human-agent interaction, would help inform agent design for positive societal impact. For instance, developers could incorporate specific priors into agents to reinforce tendencies for fair outcomes \cite{fehr1999fairness, hughes2018inequity}.

\textbf{Conclusion}\quad We proposed a method which is both effective at collaborating with humans and simple to implement. We also presented a rigorous and general methodology for evaluating with humans and eliciting their preferences. Together, these establish a strong foundation for future research on the important challenge of human-agent collaboration for benefiting society.

\section*{Acknowledgements}
The authors would like to thank Mary Cassin for creating the game sprite art; Rohin Shah, Thore Graepel, and Iason Gabriel for feedback on the draft; Lucy Campbell-Gillingham, Tina Zhu, and Saffron Huang for support in evaluating agents with humans; and Max Kleiman-Weiner, Natasha Jaques, Marc Lanctot, Mike Bowling, and Dan Roberts for useful discussions.

\section*{Funding disclosure}
This work was funded solely by DeepMind. The authors declare no competing interests.

\bibliography{main}

\newpage
\appendix

\bibliographystylea{abbrvnat}

\section{Environment details}
\label{sec:app/environment}

\subsection{Gameplay}
Players are placed in a gridworld environment containing cooking pots, tomato stations, dish stations, delivery locations, and empty counters. Players can move around and interact with these objects. By sequencing certain object interactions, players can pick up and deposit items. Each player (and each counter) can only hold one item at a time.

The objective of each episode is for the players to deliver as many tomato soup dishes as possible to delivery locations. In order to create a tomato soup dish, players must pickup tomatoes and deposit them into the cooking pot. Once there are three tomatoes in the cooking pot, it begins to cook for 20 steps. After 20 steps, the soup is fully cooked. Cooking progress for the soup is tracked by a loading bar overlaying the cooking pot. The loading bar increments over the cooking time and then turns green when the soup is ready for collection.

When the tomato soup is ready for collection, a player holding an empty dish can interact with the cooking pot to pick up the soup. The player can then deliver the soup by interacting with a delivery station while holding the completed dish. A successful delivery rewards both players and removes the dish from the game.

\subsection{Observations}
\begin{figure*}[h]
    \centering
    \begin{subfigure}{0.35\textwidth}
        \centering
        \includegraphics[height=9em]{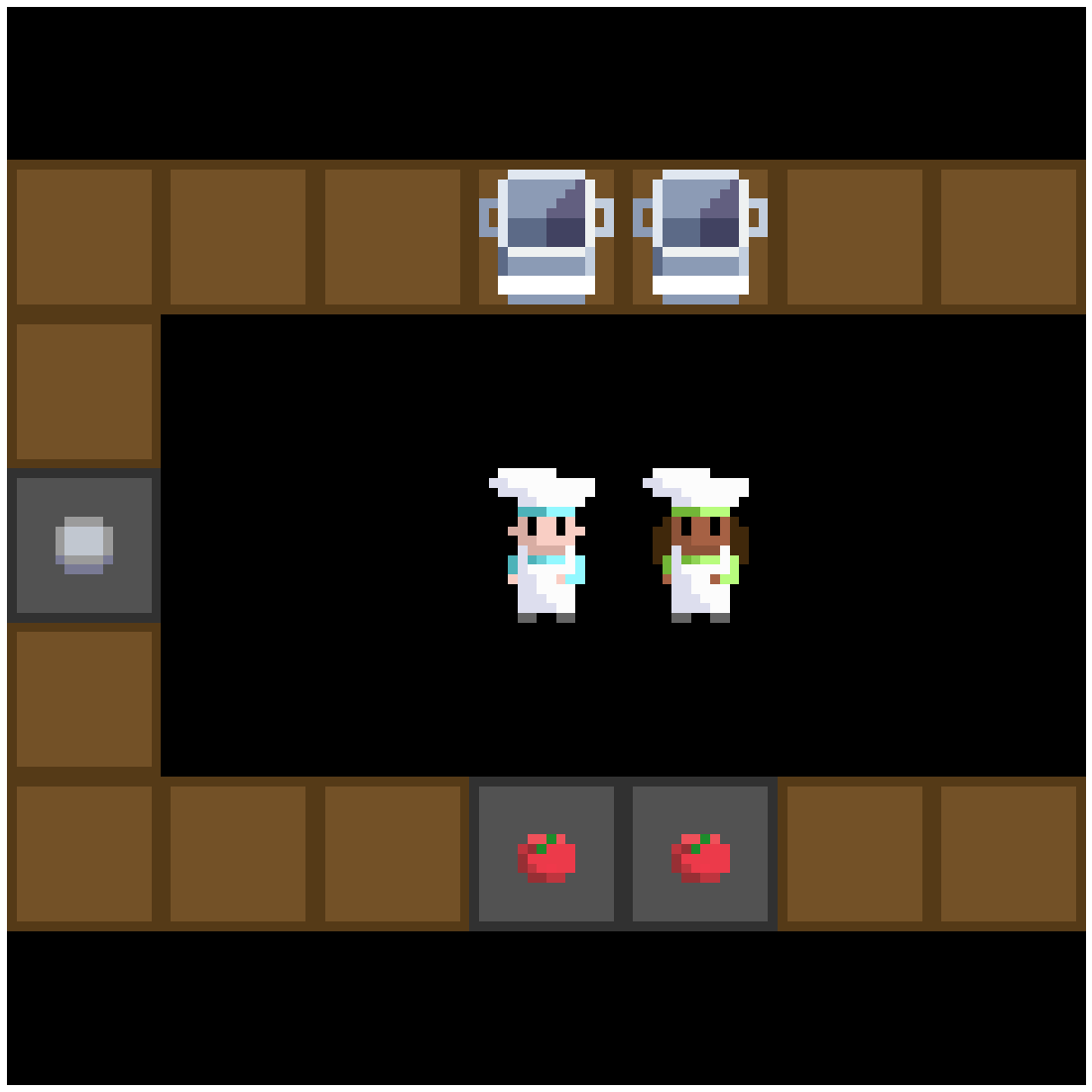}
        \caption{Agent observation (cyan).}
        \label{fig:app/environments/oc/obs/agent}
    \end{subfigure}
    \begin{subfigure}{0.35\textwidth}
        \centering
        \includegraphics[height=9em]{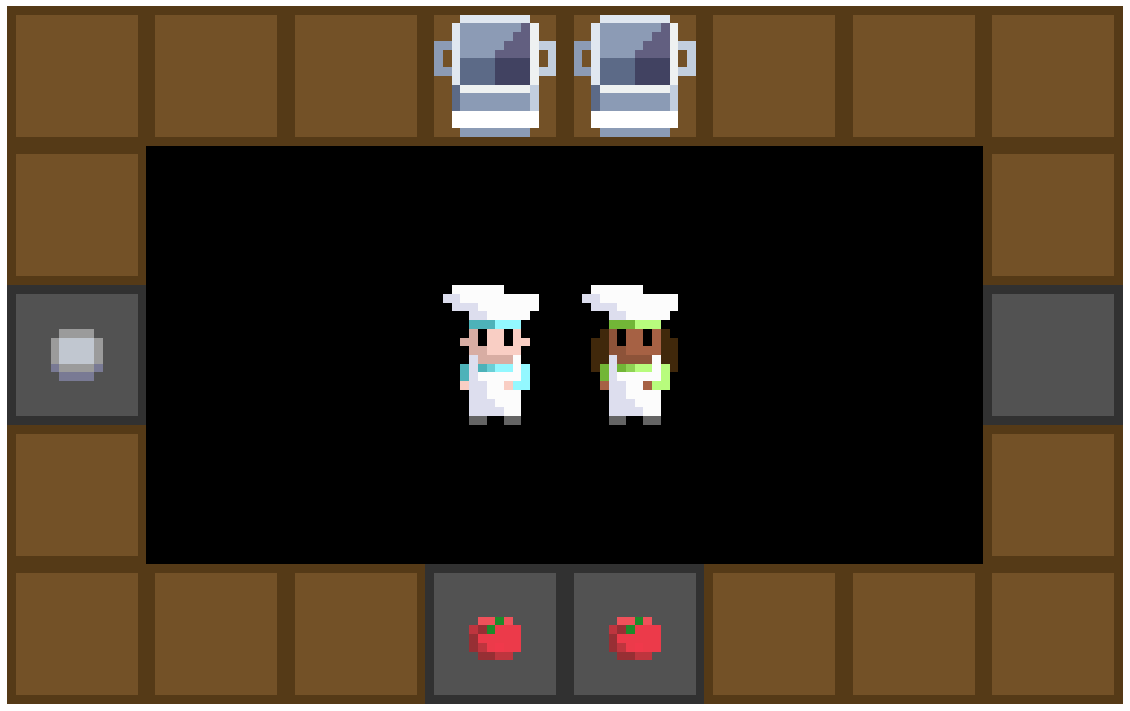}
        \caption{Human observation.}
        \label{fig:app/environments/oc/human}
    \end{subfigure}
    \caption{Example observations: (a) Agent players observe a $56\times56\times3$ egocentric view of the environment (i.e., $7\times7$ cells with a $8\times8\times3$ sprite in each cell). (b) Human players observe the full layout from a top-down perspective.}
    \label{fig:app/environments/oc/obs}
\end{figure*}

Importantly, our agent and human players observe different views of the environment. We found that agents trained better and were more robust using egocentric observations (Figure~\ref{fig:app/environments/oc/obs/agent}). However, humans players found this disorienting and so observed the world from a static top-down perspective of the environment (Figure~\ref{fig:app/environments/oc/human}).

\subsection{Actions}

Players can take one of the following eight actions each step:
\begin{enumerate}
    \item \texttt{No-op}: The player stays in the same position.
    \item \texttt{Move up}: Moves the player up one cell.
    \item \texttt{Move down}: Moves the player down one cell.
    \item \texttt{Move left}: Moves the player left one cell.
    \item \texttt{Move right}: Moves the player right one cell.
    \item \texttt{Interact}: The player interacts with the cell that they are facing.
\end{enumerate}

The outcome of the \texttt{Interact} action depends on the current item held by the player (none, empty dish, tomato, or soup), as well as the type of object which they are facing (counter, cooking pot, tomato station, dish station, or delivery station). Depending on these two conditions, the player will either deposit the held item to the object or pick up an item from the object.

\subsection{Rewards}
Players receive a shared $+20$ reward for every soup they deliver through the delivery station. As a result, they are incentivized to deliver as many soups as possible within an episode. To help scaffold learning, players also receive $+1$ reward each time they deposit a tomato into the cooking pot.

\subsection{Layouts}

\begin{figure}[h]
    \centering
     \includegraphics[width=0.94\textwidth]{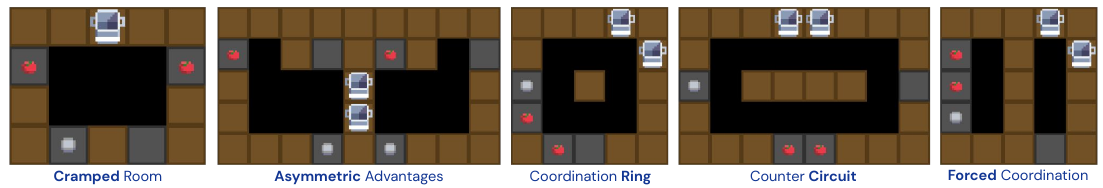}
     \caption{\textbf{Layouts:} the kitchens which agents and humans play in, each emphasizing different coordination strategies. Highlighted in bold are the terms used to refer to each in the paper.}
     \label{fig:app/environment/layouts}
\end{figure}

\begin{itemize}
    \item \textbf{Cramped:} A tight layout requiring significant movement coordination between the players in order to avoid being blocked by each other.
    \item \textbf{Asymmetric:} A two-room layout with an agent in each. In the left room, the tomato station is far away from the cooking pots while the delivery location is close. In the right room, the tomato station is next to the cooking pots while the delivery station is far. This presents an asymmetric advantage of responsibilities for optimally creating and delivering soups.
    \item \textbf{Ring:} A layout with two equally successful movement strategies -- (1) both players moving clockwise, and (2) both players moving anti-clockwise. If players do not coordinate, they will block each other's movement.
    \item \textbf{Circuit:} Players are able to cook and deliver soups by themselves through walking around the entire circuit. However, there exists a more optimal coordinated strategy whereby players pass tomatoes across the counter. Additionally, there are the clockwise and anti-clockwise strategies as in the Ring layout.
    \item \textbf{Forced:} One player is in the left room and second player is in the right room. Consequently, both players are forced to work together in order to cook and deliver soup. The player in the left room can only pass tomatoes and dishes, while the player on the right can only cook the soup and deliver it (using the items provided by the first player).
\end{itemize}

\section{Agent details}
\label{sec:app/agent}

\subsection{Fictitious Co-Play}

We provide pseudocode for Fictitious Co-Play (FCP) in Algorithm~\ref{algo:app/fcp}. Unless otherwise stated, all FCP results are with a partner population size of $N=32$. We used a checkpoint frequency $n_c$ of $1 \times 10^7$ steps, resulting in a total of 100 checkpoints saved per training run of $1 \times 10^9$ steps. After the first stage of training partners, we filter checkpoints (i.e. $F$) down to three for each partner: the first checkpoint (i.e. a randomly initialized agent), the last checkpoint (i.e. a fully trained agent), and the remaining checkpoint that achieves closest to half of the reward of the last checkpoint (i.e. a half-trained agent).

\begin{algorithm}[H]
\SetAlgoLined
\KwIn{Number of partners $N$, checkpoint frequency $n_c$, checkpoint filter $F$}
\texttt{// Stage 1: train diverse partner population} \\
partners = [] \\
\For{$i=1$ \KwTo $N$}{
 Initialize agent $i$. \\
 $n = 0$  \texttt{// step count}\\
 \While{not converged}{
  Update agent $i$ in self-play.\\
  $n \pluseq 1$ \\
  \If{$n \mod n_c = 0$}{
  Add frozen agent $i$ checkpoint to partners.
  }
 }
}
\texttt{// Stage 2: train FCP agent} \\
Filter partners with $F$. \\
Initialize FCP agent. \\
\While{not converged}{
 Sample partner from partners. \\
 Update FCP in co-play with partner. \\
}
\caption{Fictitious Co-Play (FCP)}
\label{algo:app/fcp}
\end{algorithm}

\subsection{Training settings}
Agents are trained using a distributed set of environments running in parallel. Each agent is trained using one GPU on $N \times 200$ environments, where $N$ is the number of agents being trained in the population. Agents are trained for $1 \times 10^9$ environment steps which takes between three and eight days depending on the size of the training population. As the environment involves two players, each one samples with replacement from the training population of agents every episode.

\subsection{Deep reinforcement learning (DRL)}
\subsubsection{Architecture and hyperparameters}
We used the following architecture for V-MPO \citep{song2019vmpo}. The agent's visual observations were first processed through a 3-section ResNet used in \citet{mckee2021quantifying}. Each section consisted of a convolution and $3 \times 3$ max-pooling operation (stride 2), followed by residual blocks of size 2 (i.e., a convolution followed by a ReLU nonlinearity, repeated twice,
and a skip connection from the input residual block input to the output). The entire stack was passed through one more ReLU nonlinearity. All convolutions had a kernel size of 3 and a stride of 1. The number of channels in each section was (16, 32, 32).

The resulting output was then concatenated with the previous action and reward of the agent, and processed by a single-layer MLP with 256 hidden units. This was followed by a single-layer LSTM with 256 hidden units (unrolled for 100 steps), and then a separate single-layer MLP with 256 hidden units to produce the action distribution. For the critic, we used a single-layer MLP with 256 hidden units followed by PopArt normalization (which leads to stronger multi-agent performance \citep{mckee2021quantifying}).

To train the agent, we used a discount factor of 0.99, batch sizes of 16, and the Adam optimizer (learning rate of 0.0001). We configured V-MPO with a target network update period of 100, $k = 0.5$, and an epsilon temperature of 0.1. For PopArt normalization, we used a scale lower bound of $\expnumber{1}{-2}$, an upper bound of $\expnumber{1}{6}$, and a learning rate of $\expnumber{1}{-3}$.

\subsubsection{Training curves}

\begin{figure}[h]
    \centering
    \includegraphics[scale=.5]{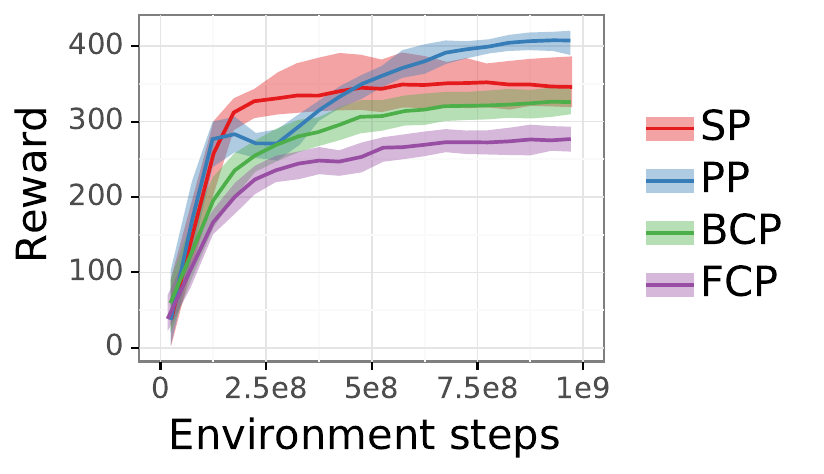}
    \caption{Training curves for agents evaluated. Median and 75\% confidence interval over 5 seeds. Note that agents vary in the partners they train with, so the scores should not be directly compared to each other (e.g. FCP training partners include  randomly initialized agents).}
    \label{fig:app/ai_ai/training_curves}
\end{figure}

\subsection{Behavioral cloning (BC)}
\subsubsection{Architecture and hyperparameters}
We trained a single network consisting of five sub-networks (1 per layout), each being a three-layer MLP with 256 hidden units per layer. The network was then trained using all of the recorded human trajectories (i.e. on all layouts), with the sub-network conditioned on the trajectory's layout feature. We used a batch size of 256 and and the Adam optimizer (learning rate = 0.0003).

\subsubsection{Feature-based observations}
As learning from RGB observations requires a significant amount of human data, we instead opted to handcraft a set of features for our BC agent to learn from. In particular, we used the following features: player position and orientation, current held item (as a one-hot vector), the relative position of the other player, the state of each cooking pot (as a one-hot vector of empty, 1 tomato, 2 tomato, 3 tomatoes, and cooked), faced cell is empty, each adjacent cell is empty (length 4), and the relative distance to each object (tomato, dish, soup, and delivery location). For our behavioral cloning, we also included the layout name.

\subsubsection{Training details}
We collected 5 human-human trajectories of length 1200 timesteps for each of the 5 layouts, resulting in 60,000 total environment steps. We then divided this data in half to train two BC agents: (1) a partner for training a BCP agent (\textit{H$_{\rm{partner}}$}), and (2) a ``human proxy'' partner for agent-agent evaluation (\textit{H$_{\rm{proxy}}$}).

\textit{H$_{\rm{partner}}$} was trained for 32,000 gradient steps (273 epochs) and \textit{H$_{\rm{proxy}}$} was trained for 61,000 gradient steps (520 epochs). Training time was based on peak reward in self-play, analogous to early stopping.

\section{Zero-shot coordination with agents}
\label{sec:app/ai_ai}

\subsection{Evaluation details}
\label{sec:app/ai_ai/evaluation_details}
For each of SP, BCP, and FCP, we trained 5 random seeds per agent. For PP, we trained a single 32-agent population and chose the first 5 agents for evaluation. All results are averaged across these seeds.

For the held-out evaluation populations, the ``diverse SP'' group consisted of 60 self-play agents varying in random seed (5 seeds), architecture (4 variants: LSTM vs feedforward, policy and value network widths of 16 vs 256), and training time (3 variants: randomly initialized, half-trained, and fully-trained). The architectural variation is the same type used for the $FCP_{+A}$ agent in Table~\ref{tab:agent_agent/results/ablation}, while the training time variation is the same used for producing partners for FCP. The random agents population were a subset of the diverse SP group, consisting of the 5 seeds of randomly initialized agents. Finally, \textit{H$_{\rm{proxy}}$} was the BC agent trained on the held-out half of human-human trajectories.

For each pair of agents consisting of one random seed of an agent to be evaluated and one random seed of an agent from a target evaluation population, we played 10 games of length $T=540$ steps for each layout.

\subsection{Additional results}
\label{sec:app/ai_ai/additional_results}

\subsubsection*{Per-layout results}

\begin{figure}[h]
     \centering
     \begin{subfigure}[b]{\textwidth}
         \centering
         \includegraphics[width=\textwidth]{figures/environment/layouts.pdf}
         \caption{Layouts for reference.}
         \label{fig:appendix/agent_agent/results/crossplay/permap/layouts}
     \end{subfigure}
     \hfill
     \begin{subfigure}[b]{\textwidth}
         \centering
         \includegraphics[width=\textwidth]{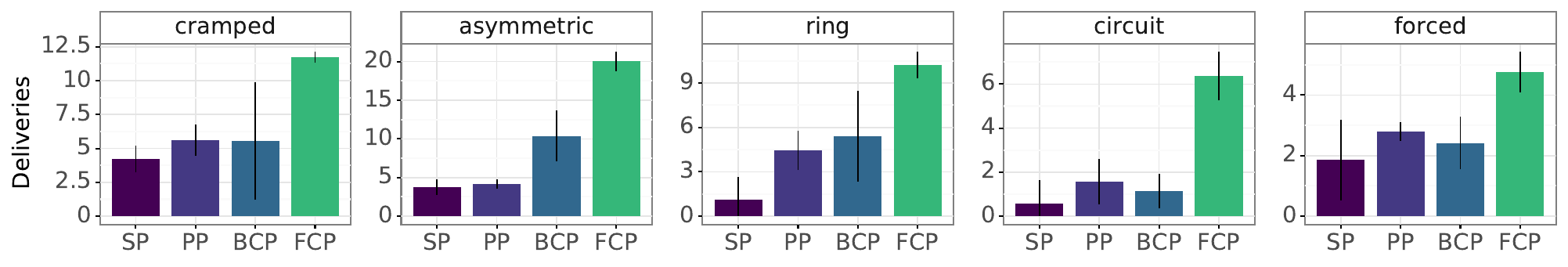}
         \caption{With \textit{H$_{\rm{proxy}}$}.}
         \label{fig:appendix/agent_agent/results/crossplay/permap/BC}
     \end{subfigure}
     \hfill
     \begin{subfigure}[b]{\textwidth}
         \centering
         \includegraphics[width=\textwidth]{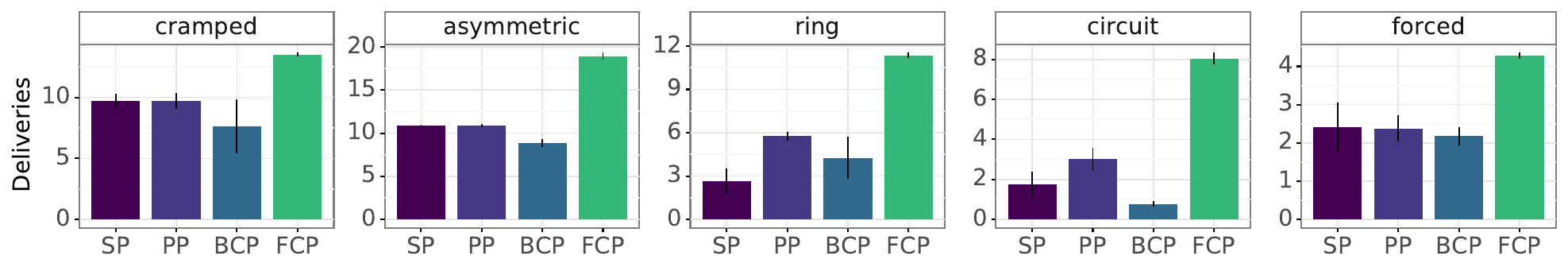}
         \caption{With diverse SP agents.}
         \label{fig:appendix/agent_agent/results/crossplay/permap/RL}
     \end{subfigure}
     \hfill
     \begin{subfigure}[b]{\textwidth}
         \centering
         \includegraphics[width=\textwidth]{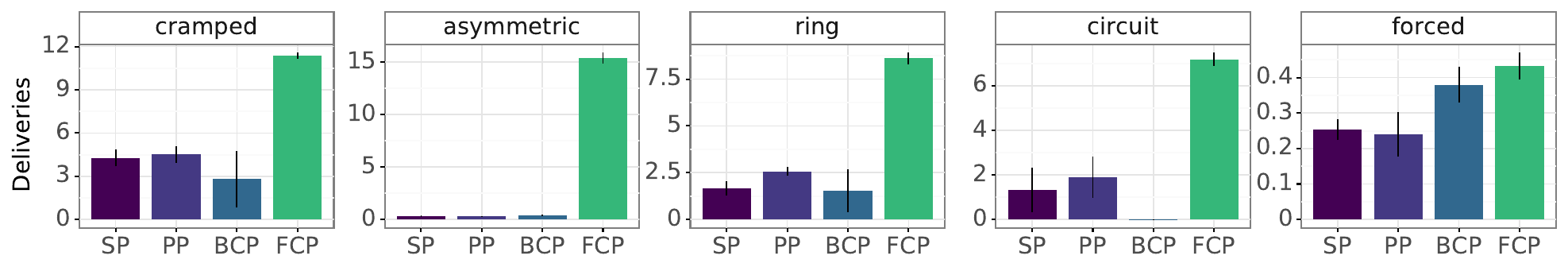}
         \caption{With random agents.}
         \label{fig:appendix/agent_agent/results/crossplay/permap/random}
     \end{subfigure}
    \caption{\textbf{Agent-agent collaborative evaluation, per-layout:} Performance of each agent when partnered with each of the held-out populations in episodes of length $T = 540$. Error bars represent standard deviation over five random training seeds. FCP outperforms all baselines on every map with every partner population.} 
    \label{fig:app/agent_agent/results/per_map}
\end{figure}

\subsubsection*{Influence of population size on performance}

As we create a training population to train our agent, a natural question to ask is how the size of that population influences the performance of the agent. In Table~\ref{tab:agent_agent/results/population_size}, we present the results of varying $N$ from $4$ to $128$.

\begin{table}[h]
  \centering
  \begin{tabular}{cc}
      \hline
       $N$ & Deliveries  \\ \hline
       4 & 8.4 (0.3) \\
       8 & 9.1 (0.4) \\
       16 & 10.2 (0.4) \\
       32 & 10.6 (0.5) \\
       64 & 10.4 (0.3) \\
       128 & 10.8 (0.6) \\ \hline           
  \end{tabular}
  \caption{Performance of FCP with \textit{H$_{\rm{proxy}}$}, as a function of number of training partners ($N$): larger populations lead to stronger agents and more deliveries. The right column presents mean deliveries with standard deviation over 5 random seeds in parentheses.}
  \label{tab:agent_agent/results/population_size}
\end{table}

We observe a consistent increase in performance as $N$ increases, plateauing around $N=32$ training partners. This supports the findings of prior work that larger populations are typically stronger, to a point \citep{knott2021tests, lowe2017multi, mckee2021quantifying}. Consequently, we selected $N=32$ as our population size across all experiments.

\section{Zero-shot coordination with humans}
\label{sec:app/human_ai}

\subsection{Experimental design}
\label{sec:app/human_ai/experimental_design}

To test how effectively the FCP and baseline agents collaborate with humans in a zero-shot setting, we recruited $N = 114$ participants from Prolific, an online participant recruitment platform \citep{eyal2021data, peer2017beyond}. Inclusion criteria were residence in the United States, a minimum approval rate of 95\% on prior Prolific studies, and a minimum of 20 prior approved studies.

The study consisted of multiple tutorial pages explaining the game rules and dynamics (Figures~\ref{fig:app/screenshots_1} and \ref{fig:app/screenshots_2}), a single-player practice episode (Figure~\ref{fig:app/screenshots_3}), multiple two-player episodes with agent partners (Figures~\ref{fig:app/screenshots_4}-\ref{fig:app/screenshots_6}), and a debrief questionnaire collecting open-ended feedback and demographic information. Each participant played multiple two-player episodes on different layouts. By the end of the study, each participant had collaborated with agents generated through every training method (i.e., the study had a within-participant design). We incentivized game performance: participants earned \$0.10 for each dish served, resulting in a cumulative performance bonus at the end of the study. Study sessions lasted 31.2 minutes on average, with a compensation base of \$4.00 and an average bonus of \$6.74.

DeepMind's independent research ethics committee conducted ethical review for the project and offered a favorable opinion for the study protocol (\#19/001), indicating that it presented minimal risk to participants.
All participants provided informed consent for the study.

Each participant played with a randomized sequence of agent partners and game layouts. Specifically, we first randomized the order of the five layouts. Layouts were repeated four times in the sequence, so that each participant played 20 episodes total (four episodes on each layout). To generate the sequence of agents that plays in the four episodes for each layout, we sampled four agents without replacement.

\clearpage
\subsubsection{Screenshots}
\label{sec:app/human_ai/screenshots}

\begin{figure*}[h]
    \captionsetup[subfigure]{labelformat=empty}
    \centering
    \begin{subfigure}{0.32\textwidth}
        \centering
        \includegraphics[width=\linewidth]{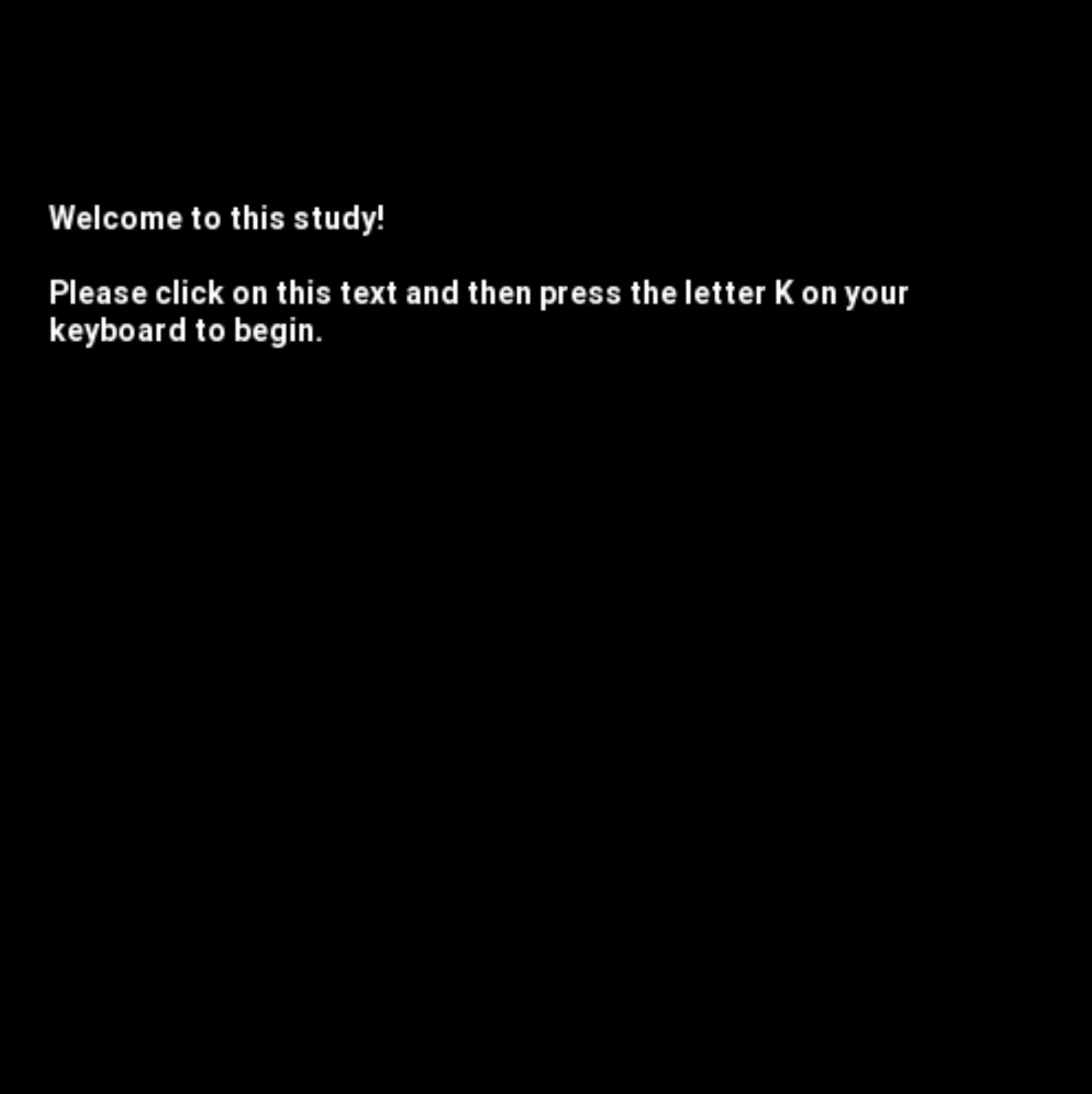}
        \caption{Screen 1: Welcome participants to the experiment.}
    \end{subfigure}
    \hfill
    \begin{subfigure}{0.32\textwidth}
        \centering
        \includegraphics[width=\linewidth]{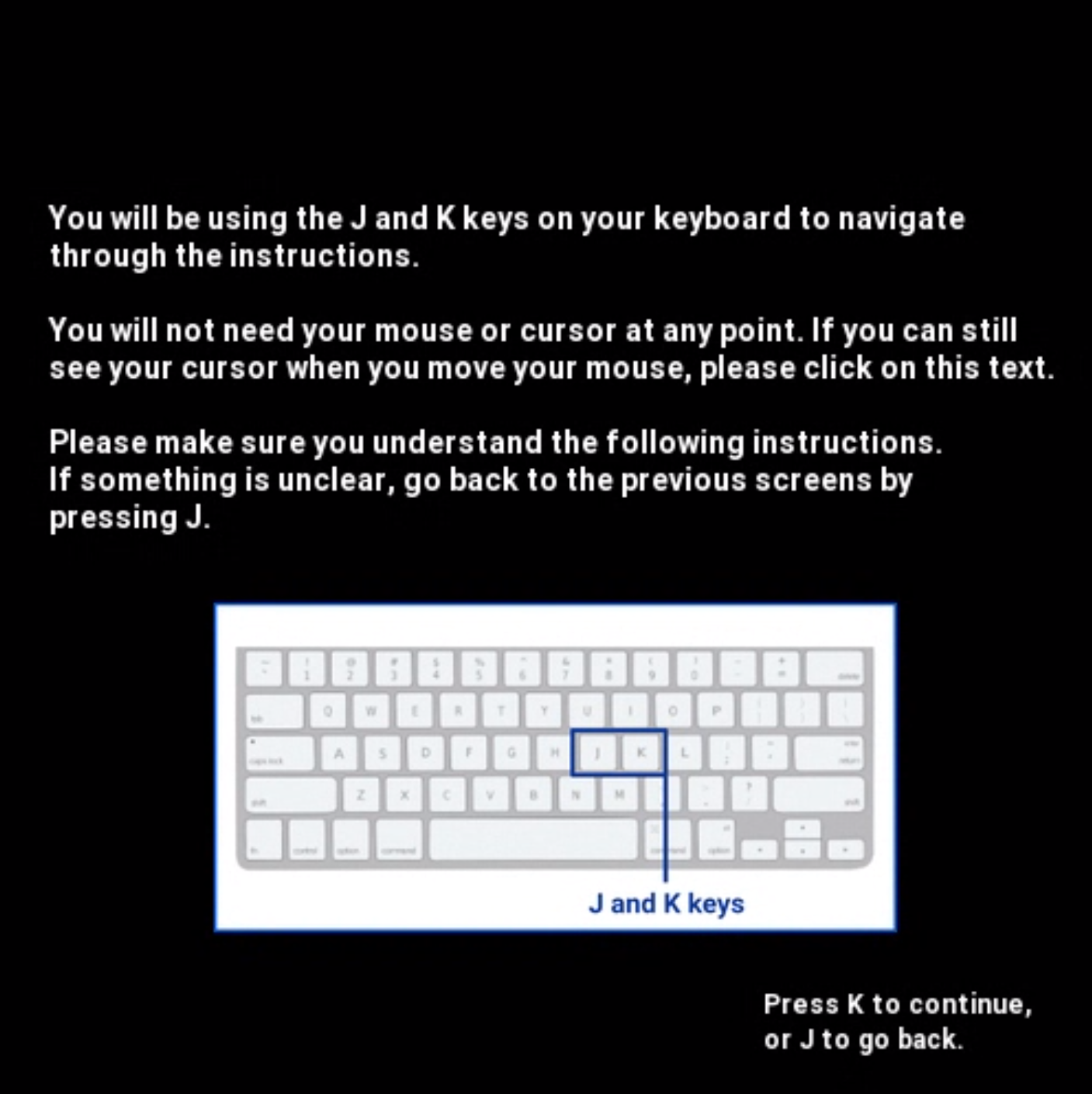}
        \caption{Screen 2: Explain the keyboard \\ controls.}
    \end{subfigure}
    \hfill
    \begin{subfigure}{0.32\textwidth}
        \centering
        \includegraphics[width=\linewidth]{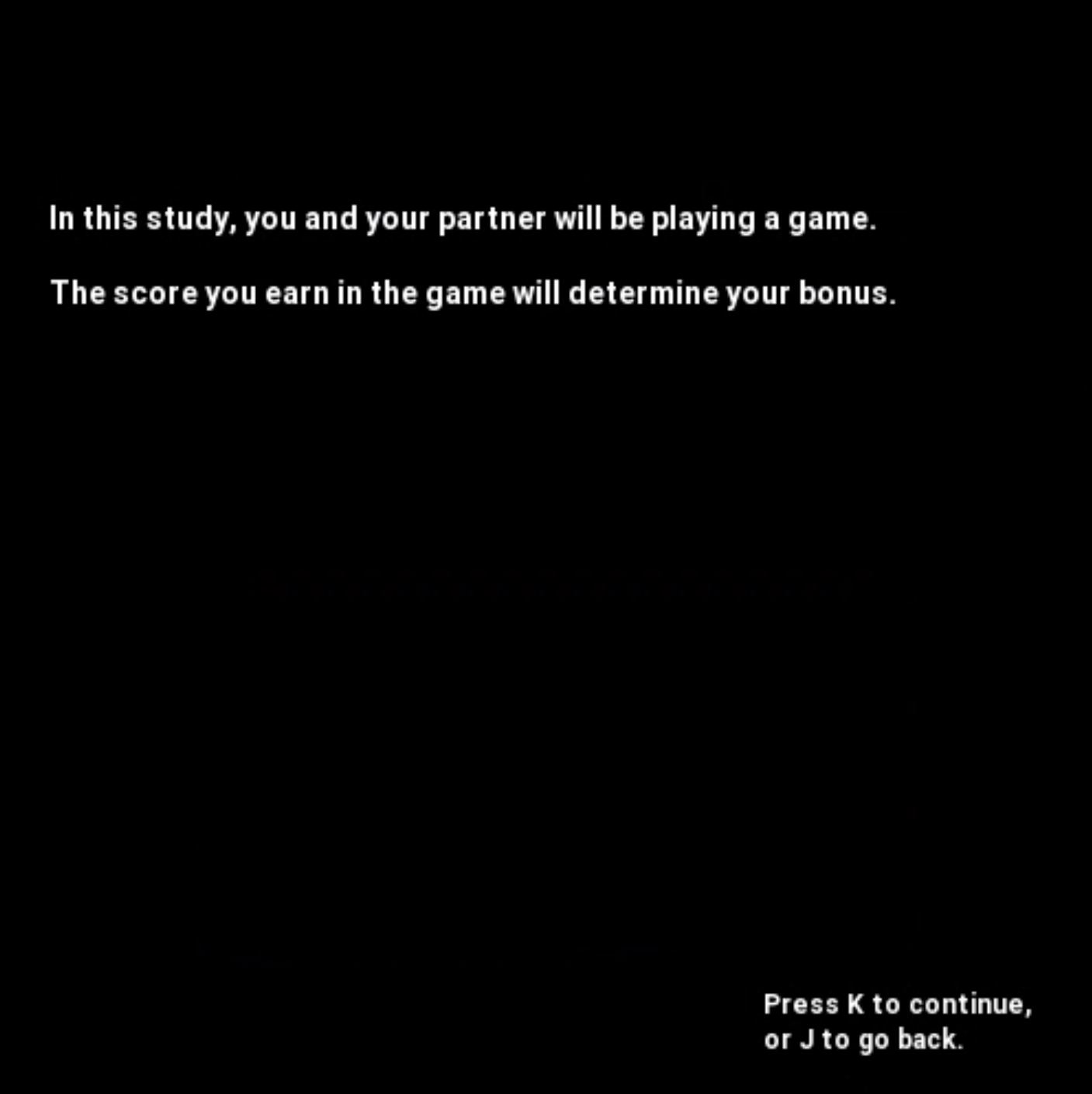}
        \caption{Screen 3: Provide overview of the \\ experiment and bonus.}
    \end{subfigure}
    \begin{subfigure}{0.45\textwidth}
        \centering
        \includegraphics[width=\linewidth]{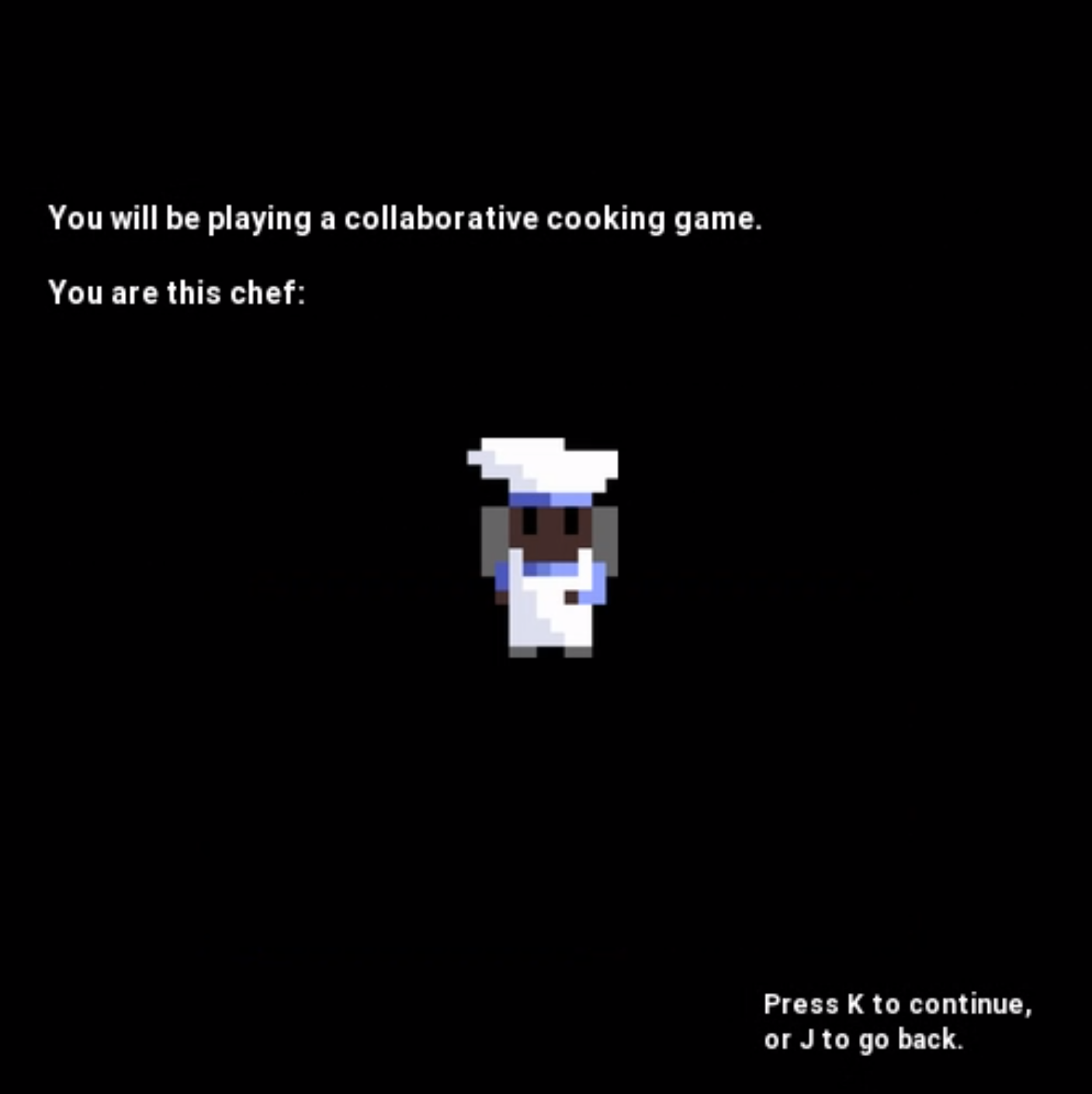}
        \caption{Screen 4: Introduce the participant's \\ controllable chef.}
    \end{subfigure}
    \hfill
    \begin{subfigure}{0.45\textwidth}
        \centering
        \includegraphics[width=\linewidth]{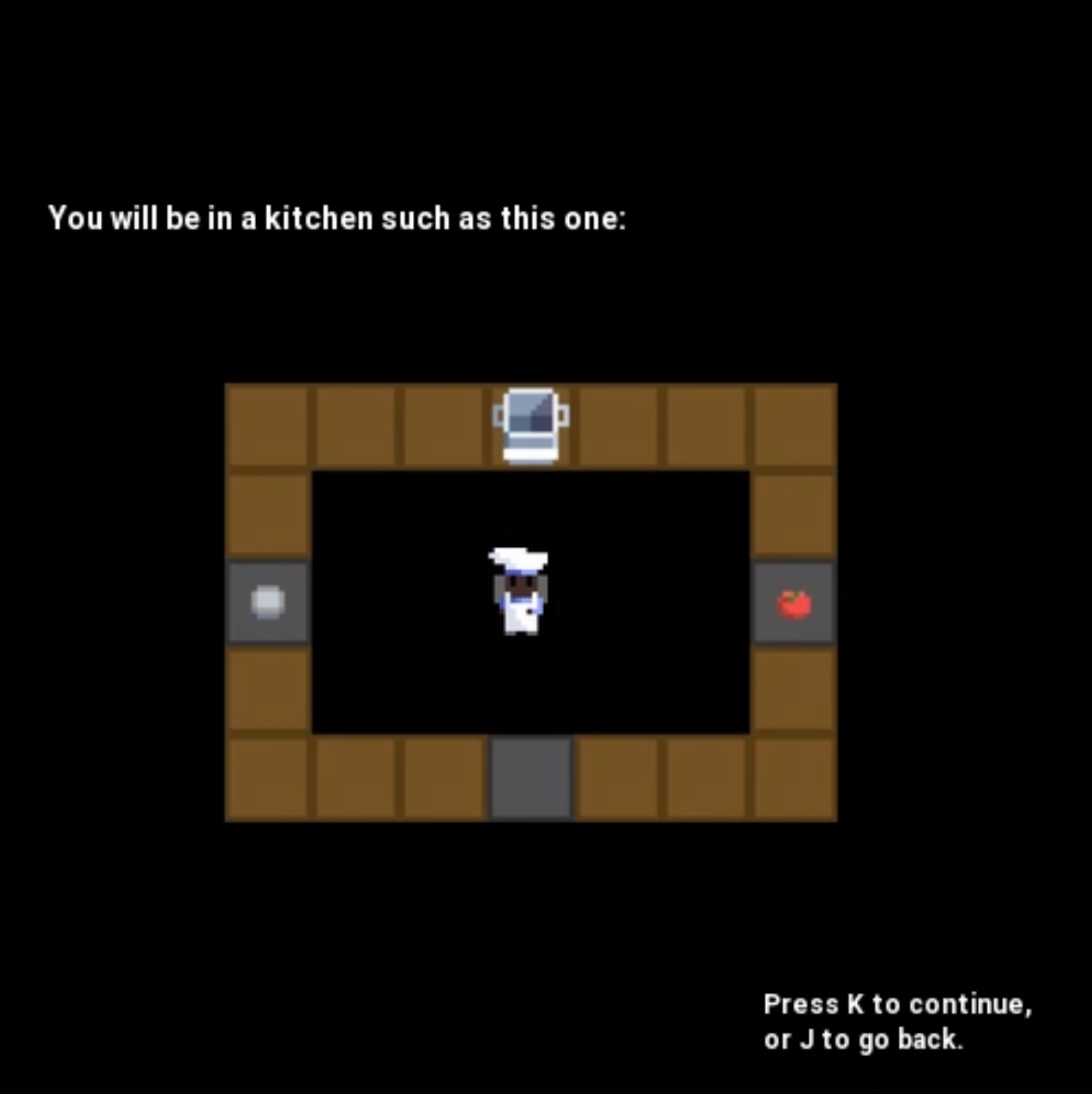}
        \caption{Screen 5: Introduce a generic kitchen layout.}
    \end{subfigure}
    \caption{Screenshots of tutorial and instruction screens.}
    \label{fig:app/screenshots_1}
\end{figure*}

\begin{figure*}[h]
    \captionsetup[subfigure]{labelformat=empty}
    \centering
    \begin{subfigure}{0.45\textwidth}
        \centering
        \includegraphics[width=\linewidth]{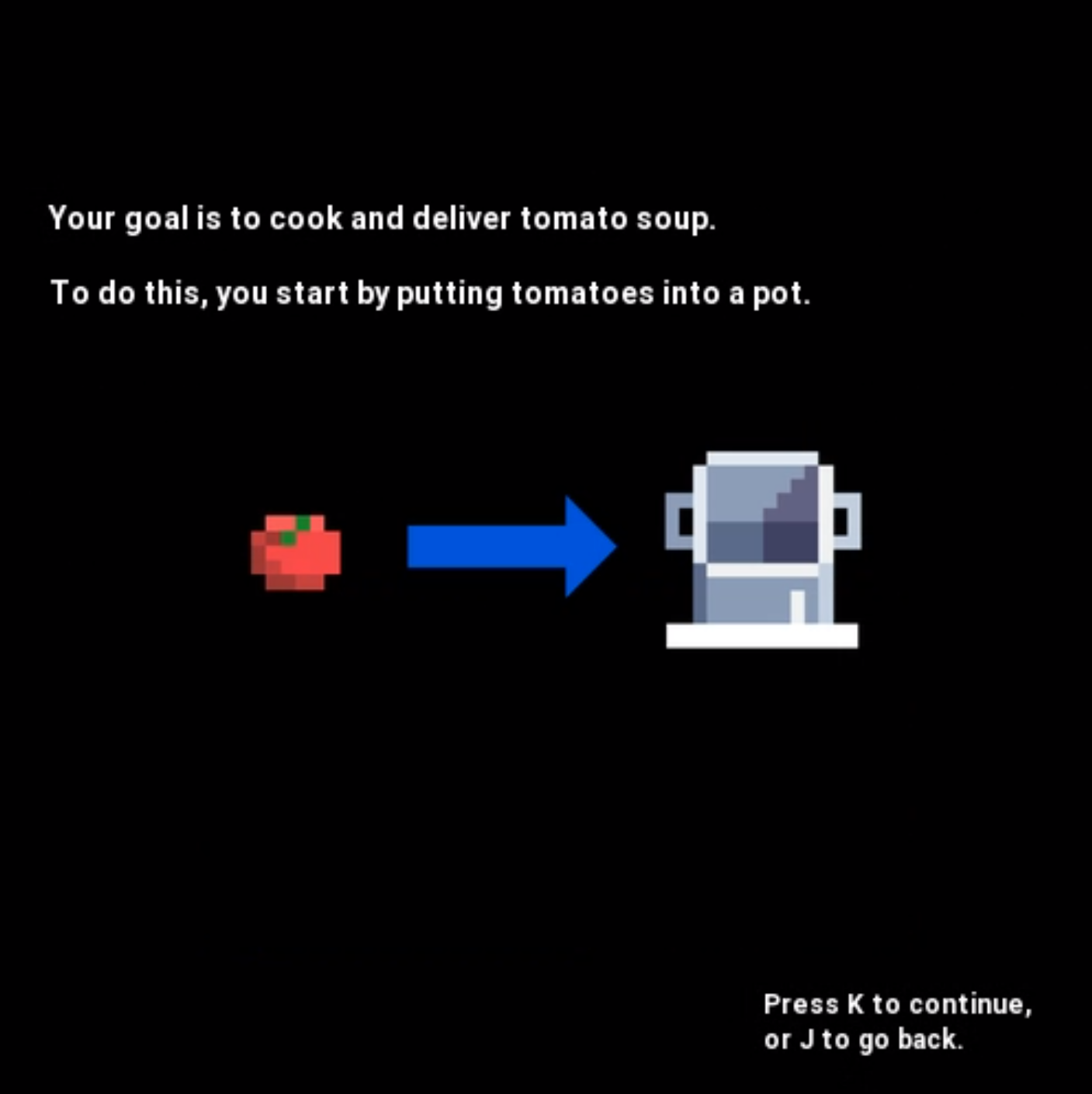}
        \caption{Screen 6: Explain the goal of the game.}
    \end{subfigure}    
    \hfill    
    \begin{subfigure}{0.45\textwidth}
        \centering
        \includegraphics[width=\linewidth]{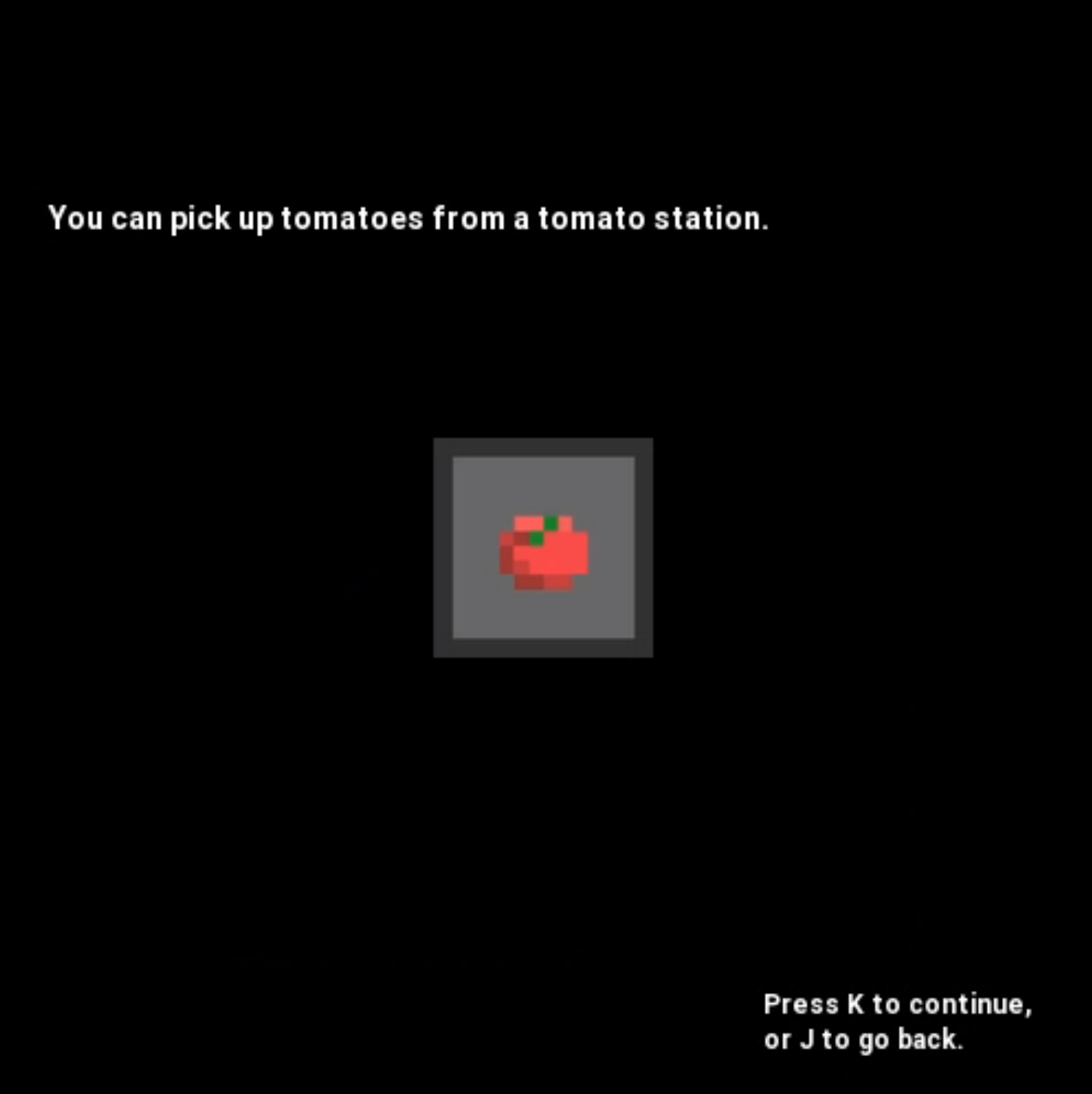}
        \caption{Screen 7: Explain how to collect tomatoes.}
    \end{subfigure}
    \hfill
    \begin{subfigure}{0.45\textwidth}
        \centering
        \includegraphics[width=\linewidth]{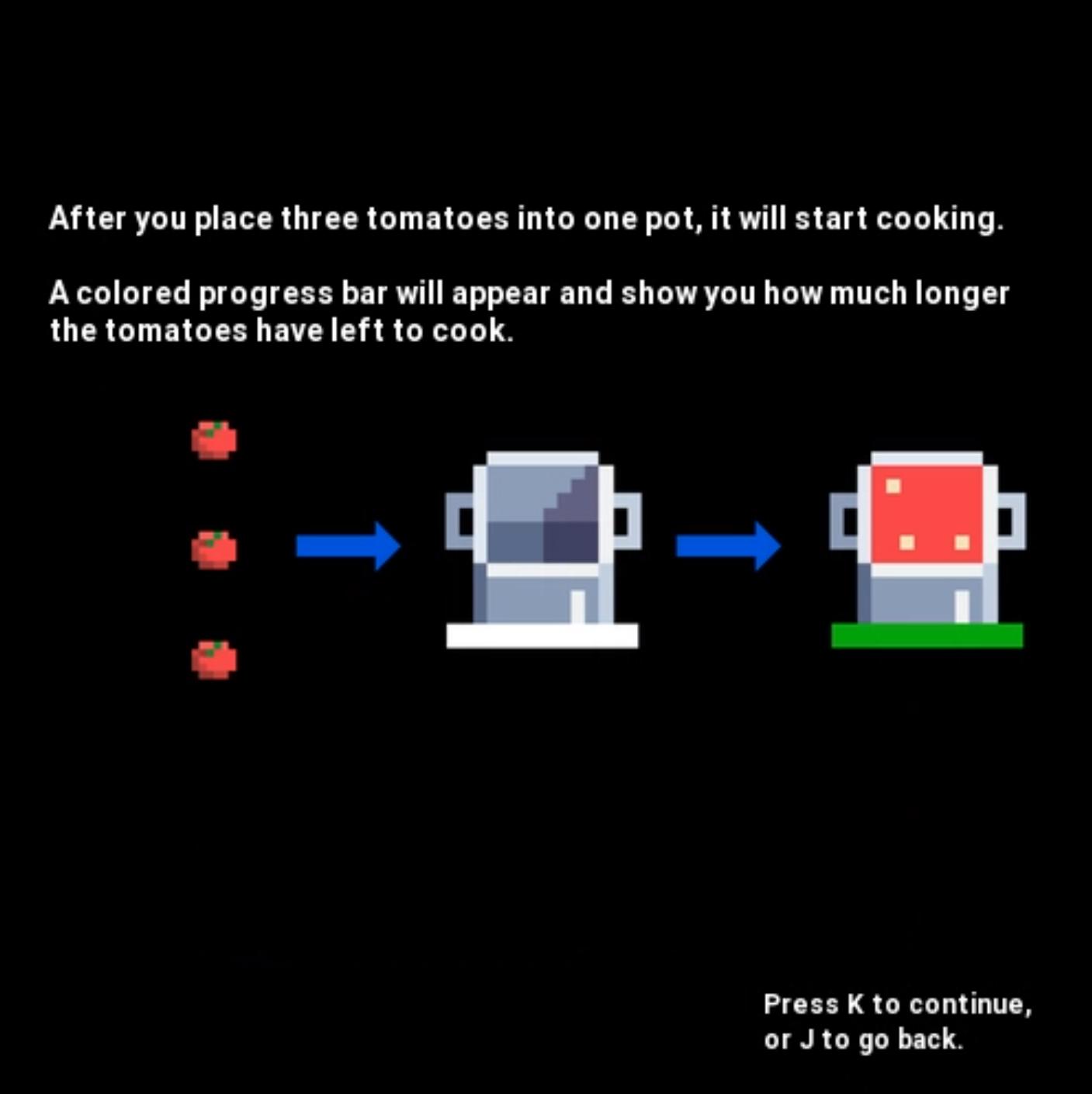}
        \caption{Screen 8: Explain how to cook soup.}
    \end{subfigure}
    \hfill
    \begin{subfigure}{0.45\textwidth}
        \centering
        \includegraphics[width=\linewidth]{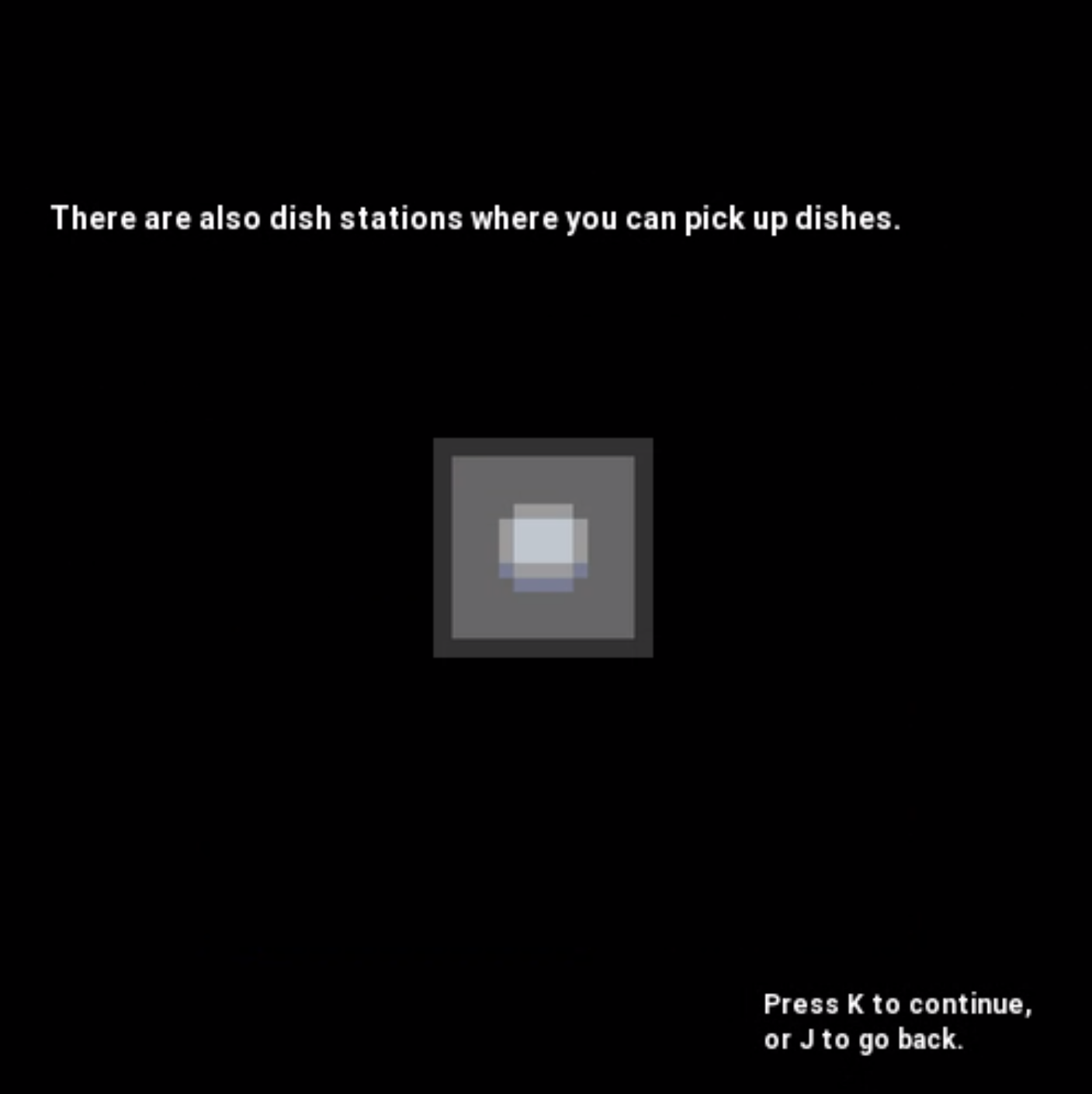}
        \caption{Screen 9: Explain how to collect dishes.}
    \end{subfigure}
    \hfill
    \begin{subfigure}{0.32\textwidth}
        \centering
        \includegraphics[width=\linewidth]{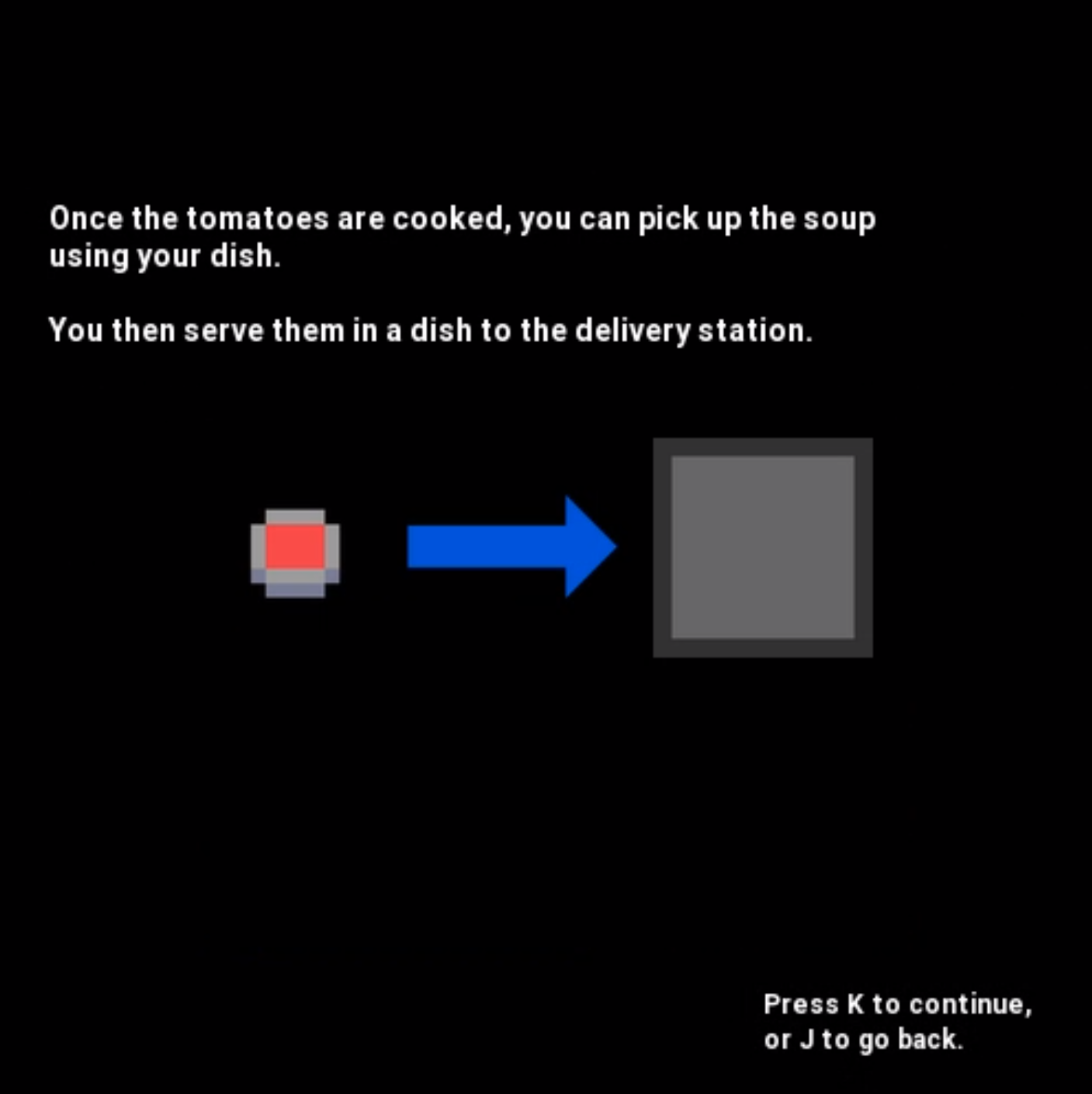}
        \caption{Screen 10: Explain how to deliver soup.}
    \end{subfigure}
    \hfill
    \begin{subfigure}{0.32\textwidth}
        \centering
        \includegraphics[width=\linewidth]{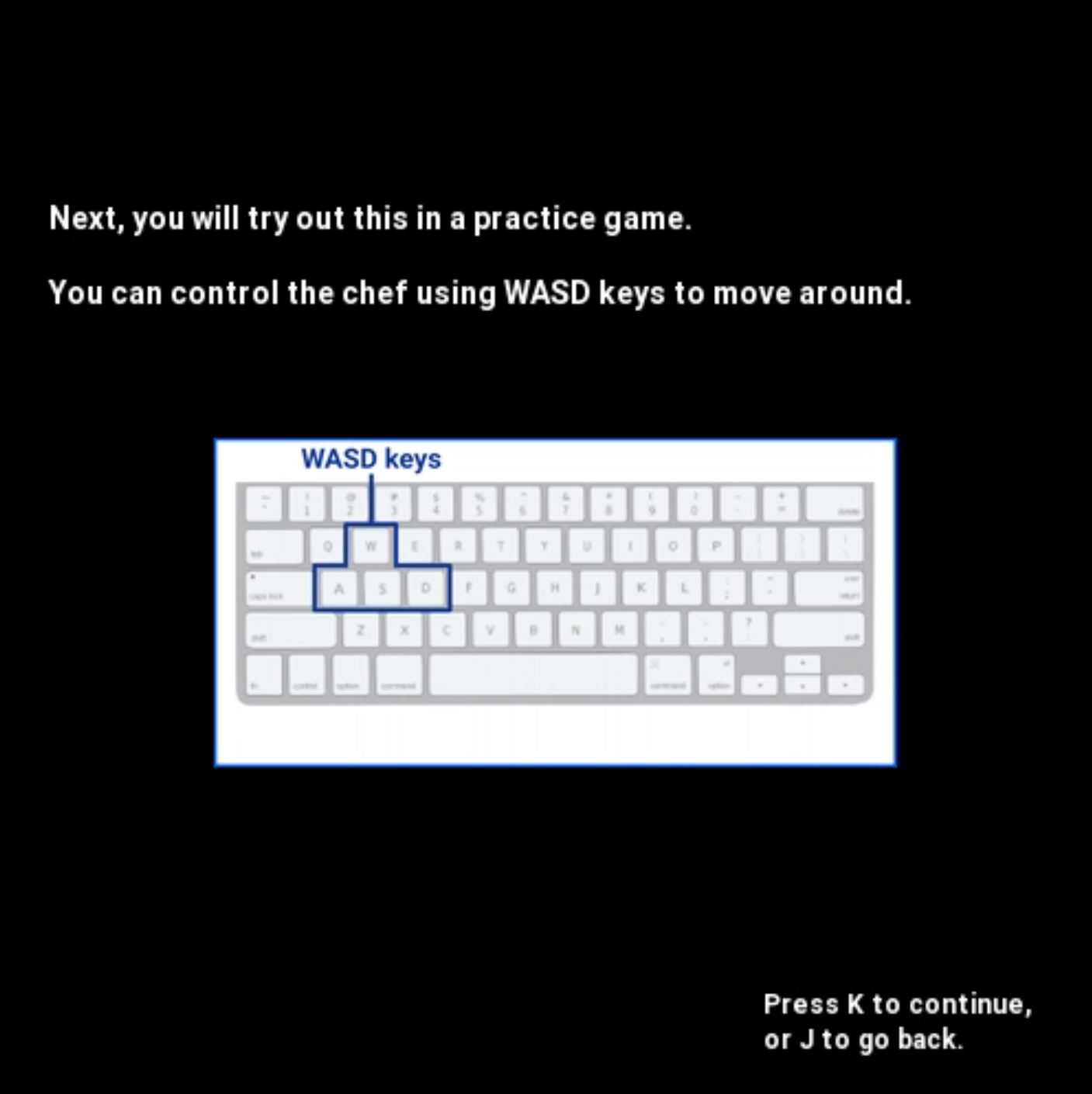}
        \caption{Screen 11: Explain the movement controls for the chef.}
    \end{subfigure}
    \hfill
    \begin{subfigure}{0.32\textwidth}
        \centering
        \includegraphics[width=\linewidth]{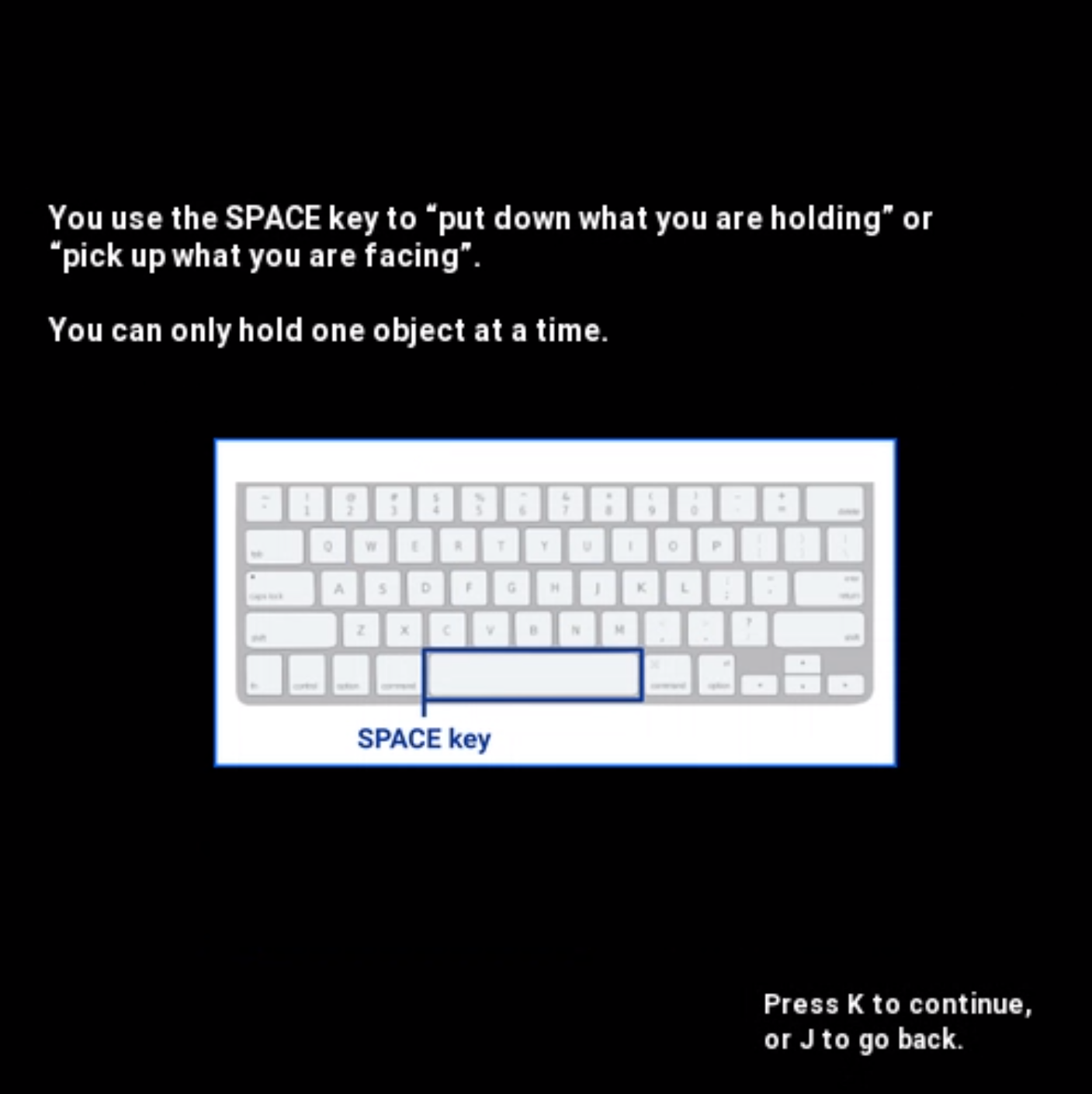}
        \caption{Screen 12: Explain the chef's \texttt{interact} action.}
    \end{subfigure}    
    \caption{Screenshots of tutorial and instruction screens.}
    \label{fig:app/screenshots_2}
\end{figure*}

\begin{figure*}
    \centering
    \begin{subfigure}{0.45\textwidth}
        \centering
        \includegraphics[width=\linewidth]{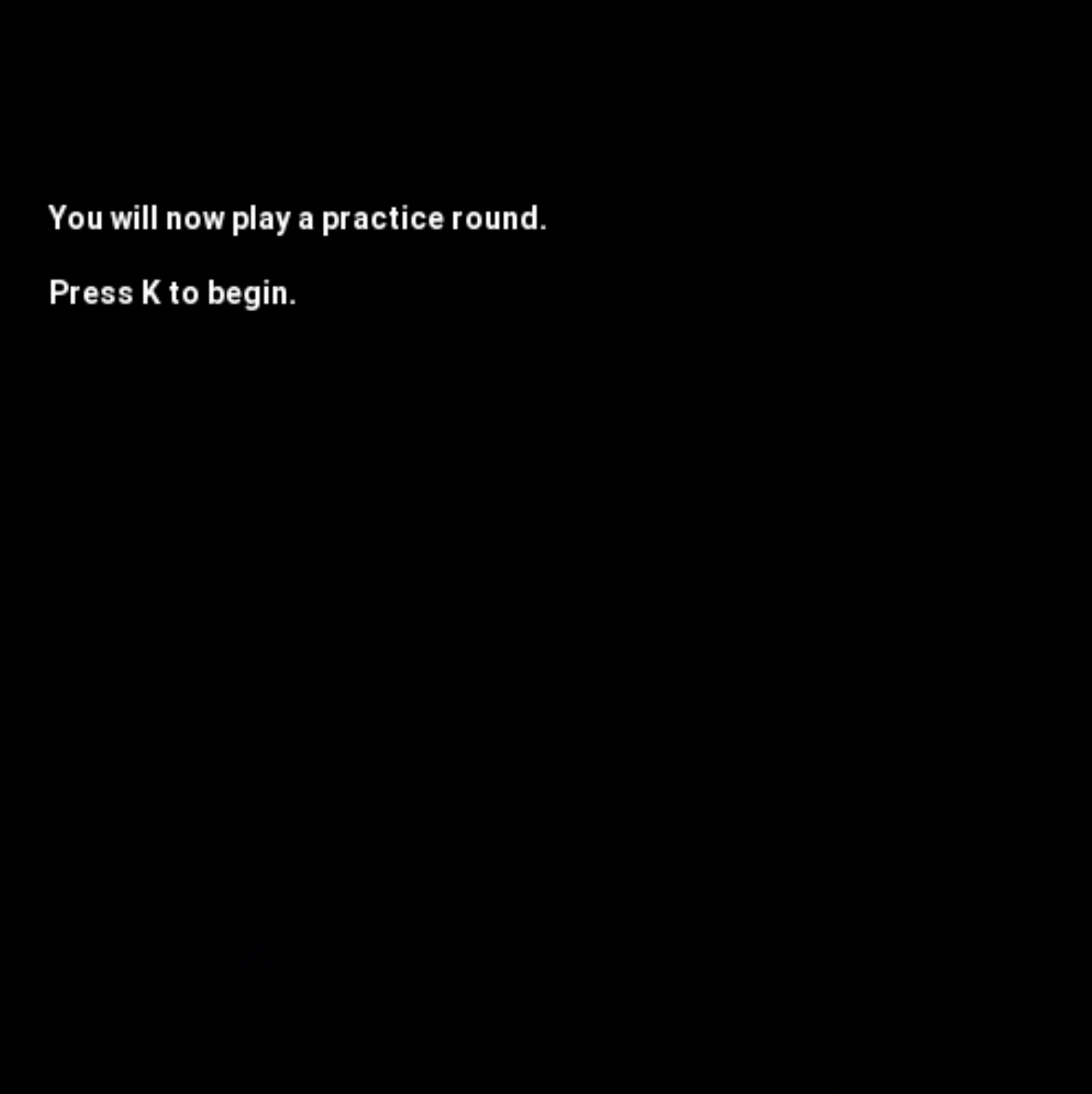}
        \caption{Screen 13: Introduce the practice episode.}
    \end{subfigure}
    \hfill
    \begin{subfigure}{0.45\textwidth}
        \centering
        \includegraphics[width=\linewidth]{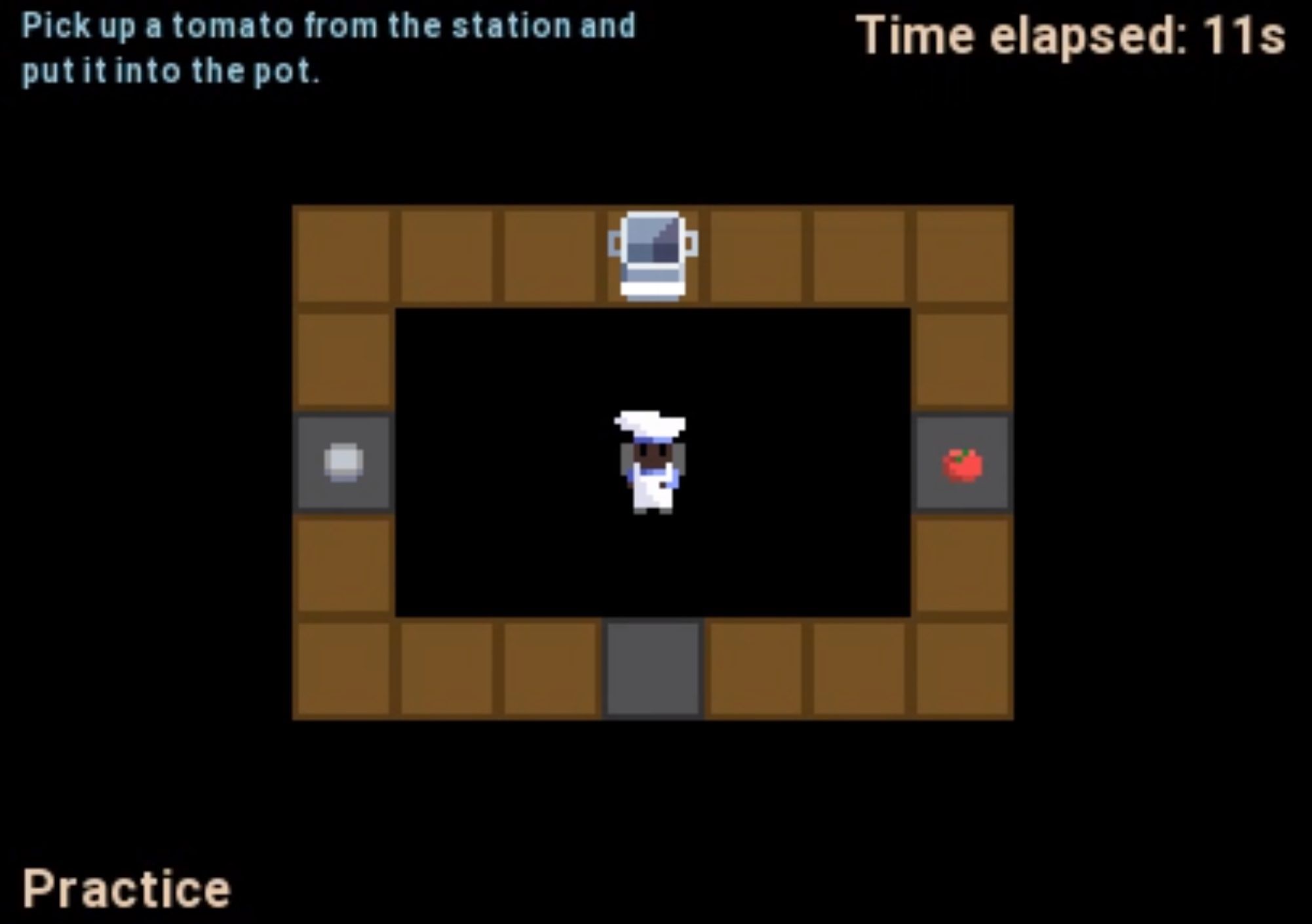}
        \caption{Screen 14: First part of the practice episode.}
    \end{subfigure}
    \hfill
    \begin{subfigure}{0.45\textwidth}
        \centering
        \includegraphics[width=\linewidth]{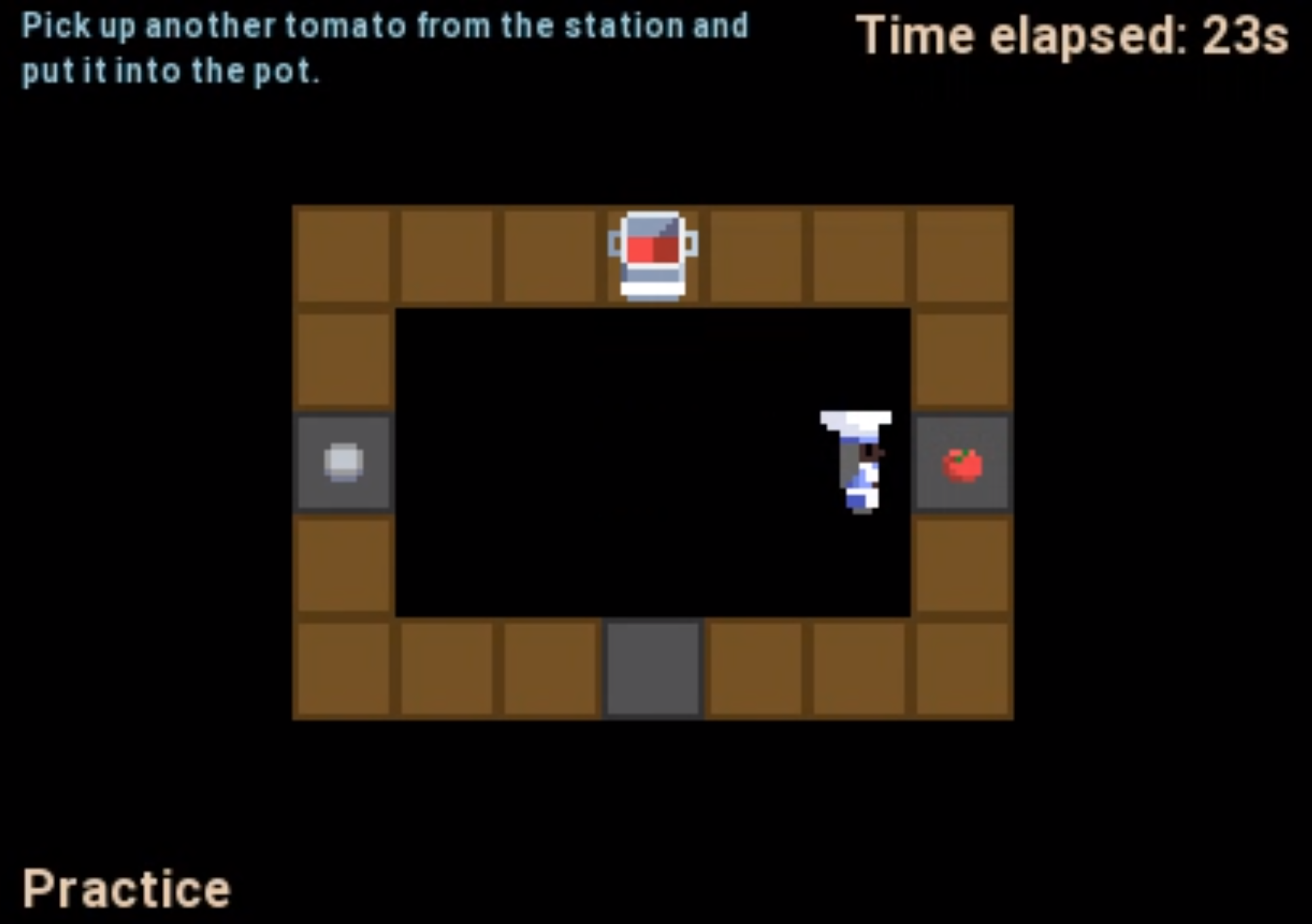}
        \caption{Screen 14: Second part of the practice episode, after the participant has placed one tomato into the cooking pot.}
    \end{subfigure}    
    \hfill
    \begin{subfigure}{0.45\textwidth}
        \centering
        \includegraphics[width=\linewidth]{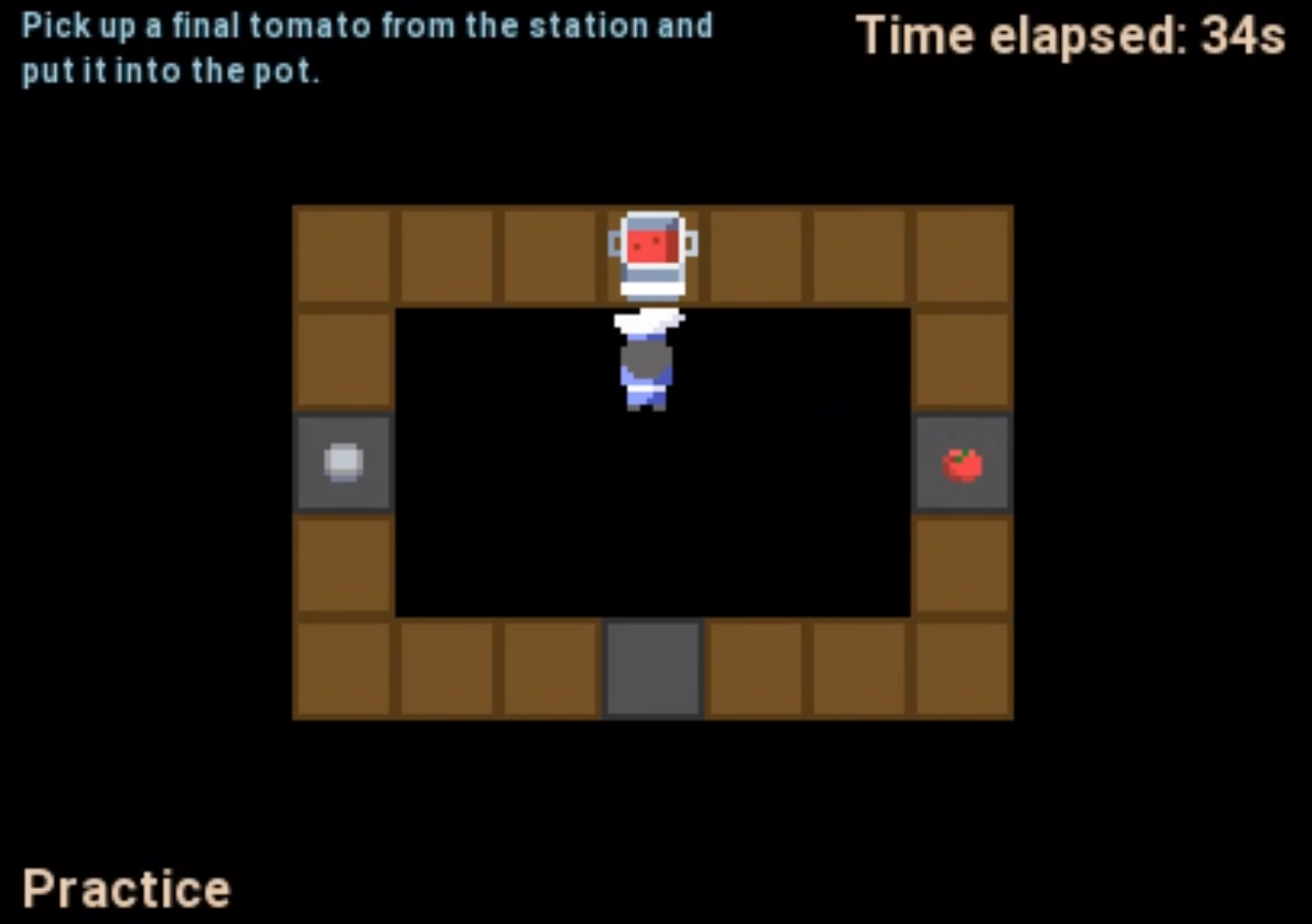}
        \caption{Screen 14: Third part of the practice episode, after the participant has placed two tomatoes into the cooking pot.}
    \end{subfigure}
    \hfill
    \begin{subfigure}{0.45\textwidth}
        \centering
        \includegraphics[width=\linewidth]{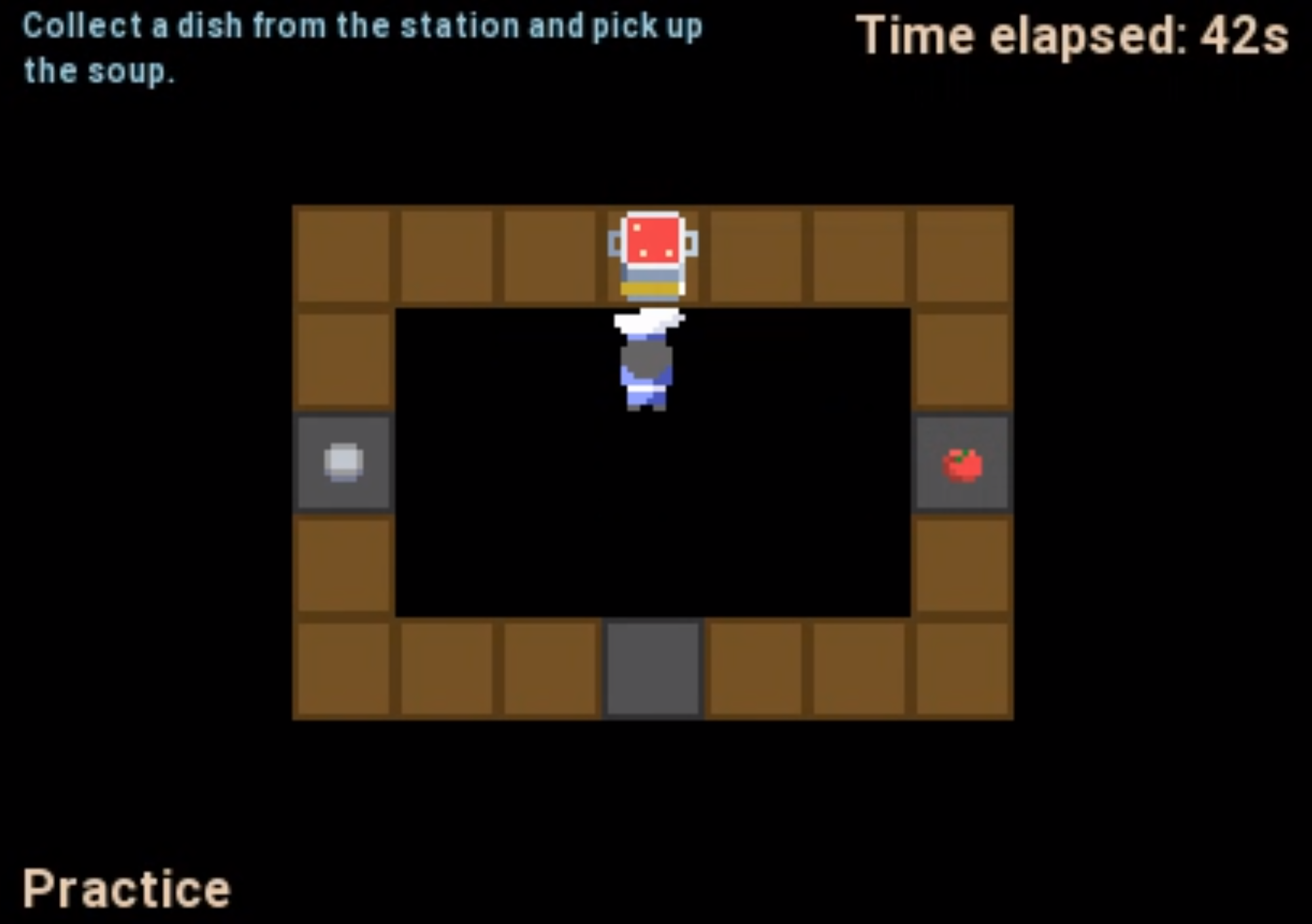}
        \caption{Screen 14: Fourth part of the practice episode, after the participant has placed three tomatoes into the cooking pot.}
    \end{subfigure}
    \hfill
    \begin{subfigure}{0.45\textwidth}
        \centering
        \includegraphics[width=\linewidth]{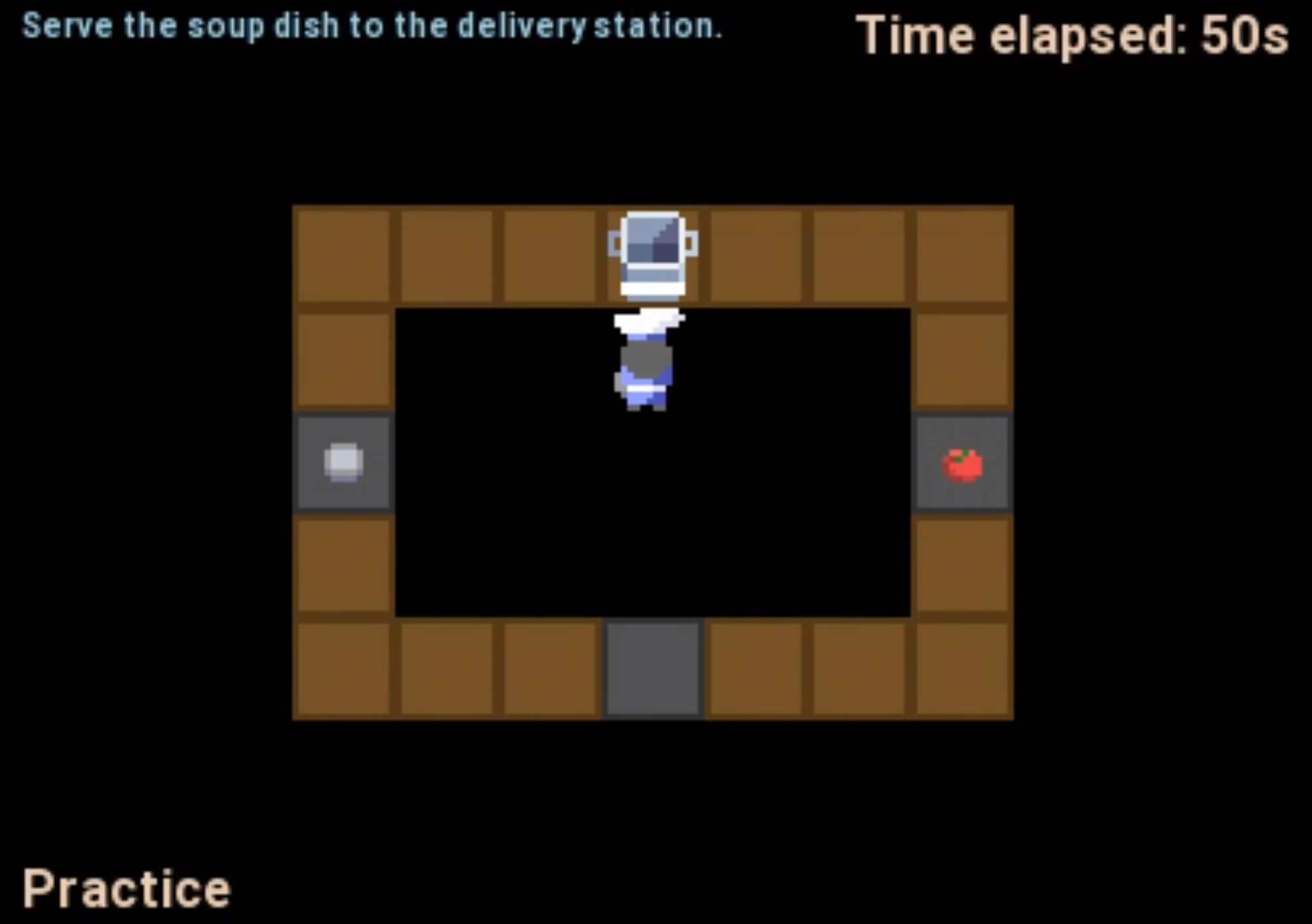}
        \caption{Screen 14: Final part of the practice episode, after the participant has collected the soup from the cooking pot.}
    \end{subfigure}
    \caption{Screenshots of the practice episode.}
    \label{fig:app/screenshots_3}
\end{figure*}

\begin{figure*}
    \centering
    \begin{subfigure}{0.32\textwidth}
        \centering
        \includegraphics[width=\linewidth]{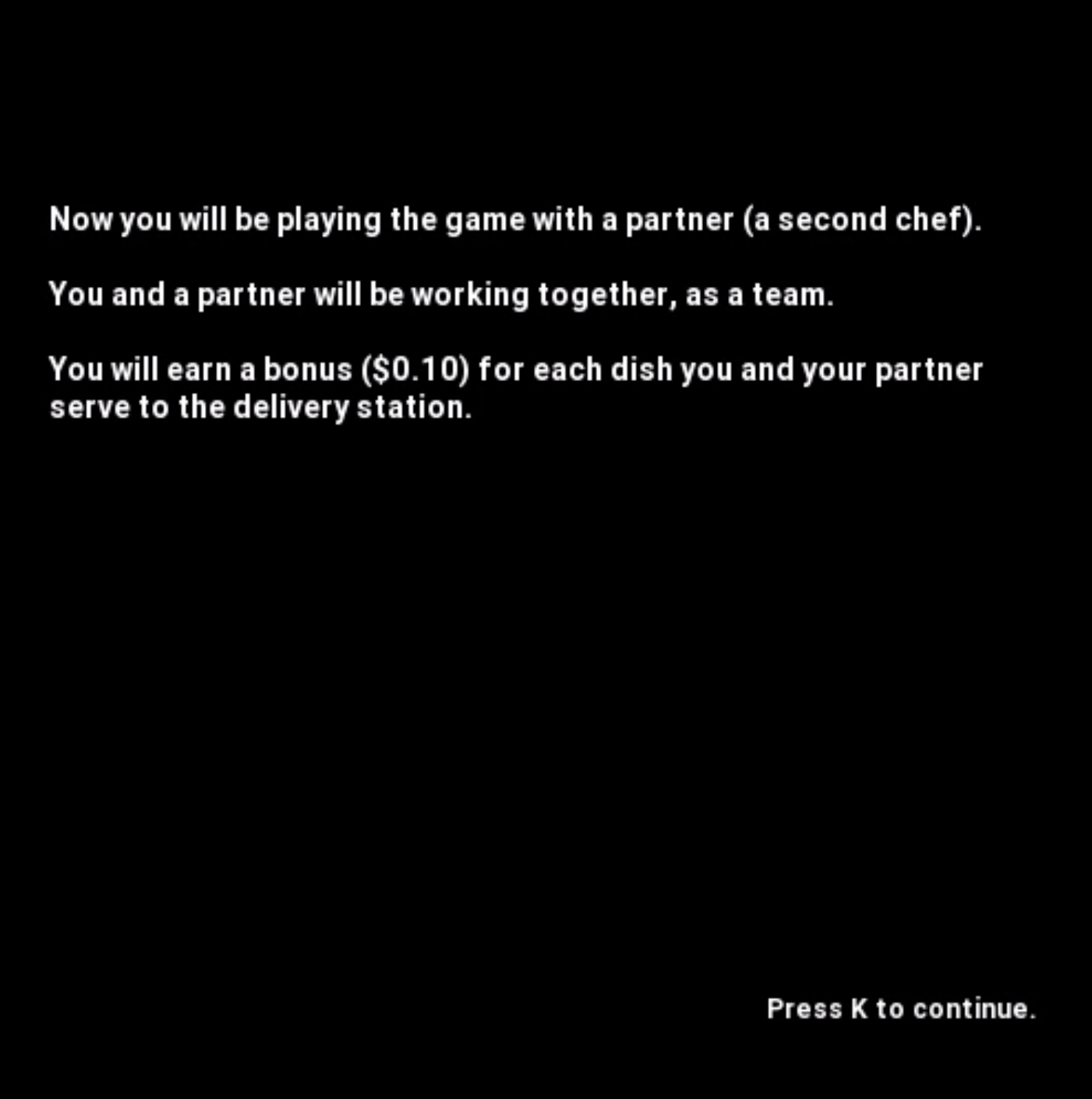}
        \caption{Screen 15: Explain the payment structure for the experiment.}
    \end{subfigure}
    \hfill
    \begin{subfigure}{0.32\textwidth}
        \centering
        \includegraphics[width=\linewidth]{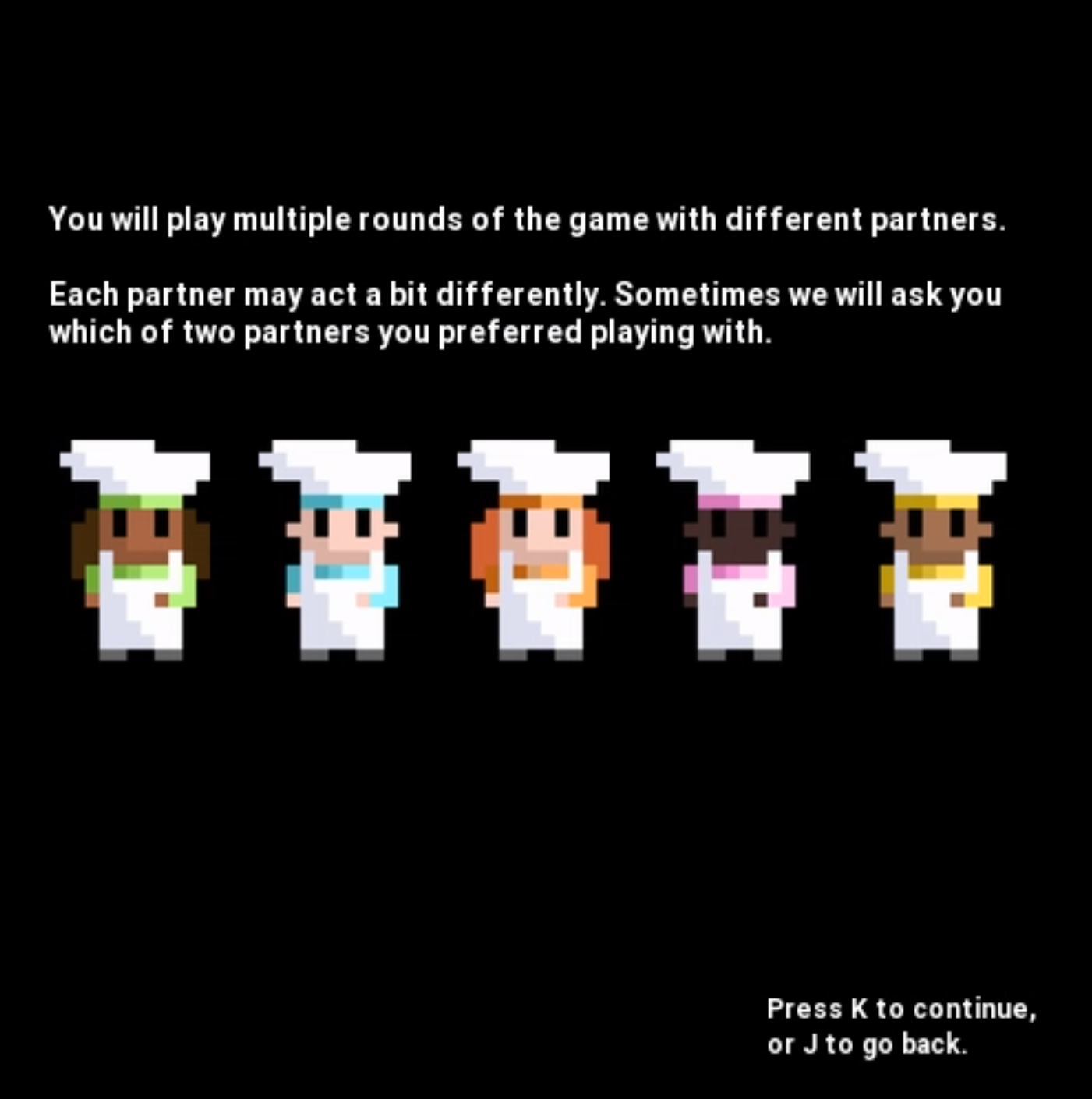}
        \caption{Screen 16: Introduce the participants' partners.}
    \end{subfigure}
    \hfill
    \begin{subfigure}{0.32\textwidth}
        \centering
        \includegraphics[width=\linewidth]{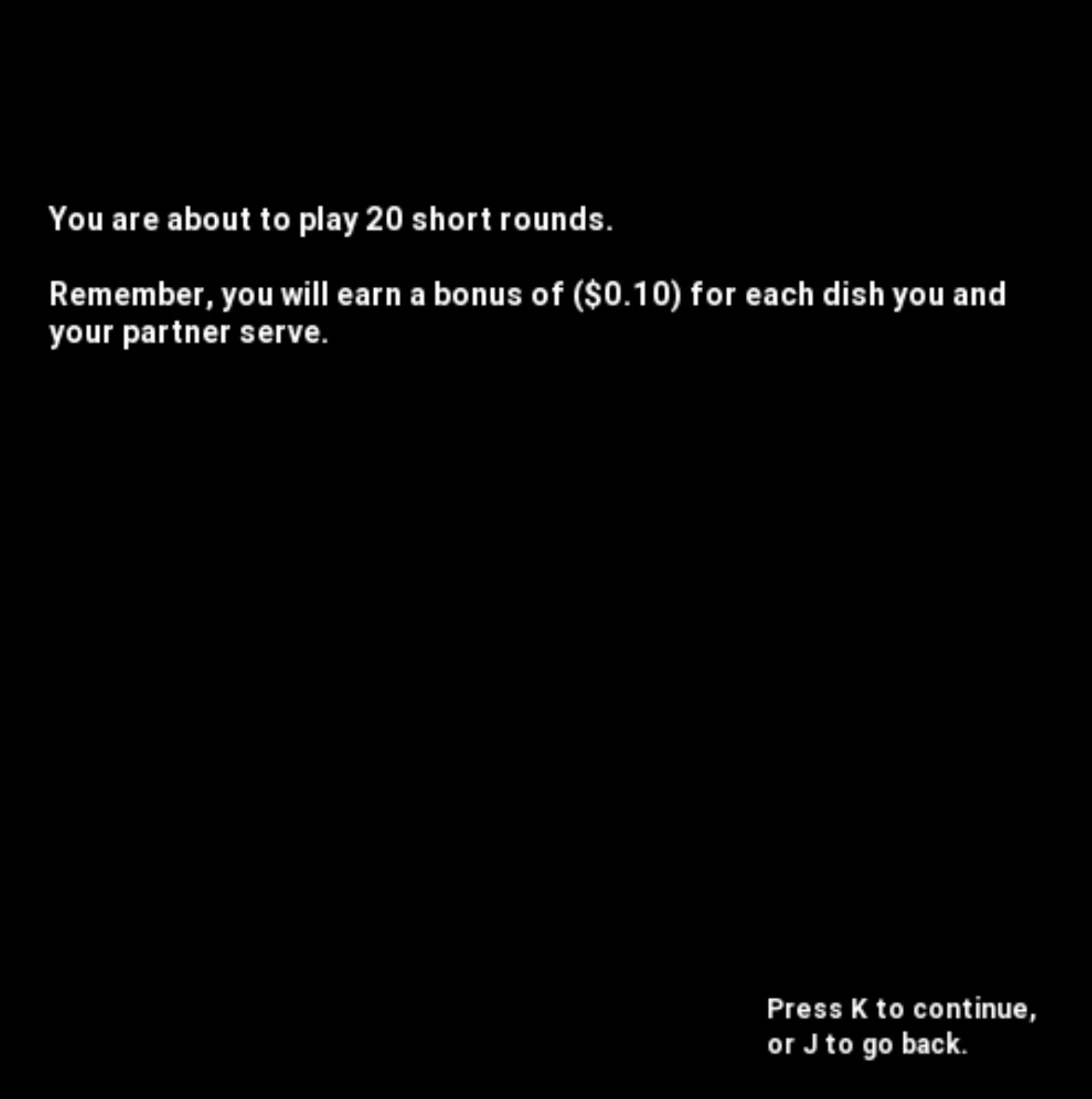}
        \caption{Screen 17: Provide overview of upcoming episodes.}
    \end{subfigure}    
    \caption{Screenshots of instruction pages.}
    \label{fig:app/screenshots_4}
\end{figure*}

\begin{figure*}[h]
    \captionsetup[subfigure]{labelformat=empty}
    \centering
    \begin{subfigure}{0.45\textwidth}
        \centering
        \includegraphics[width=\linewidth]{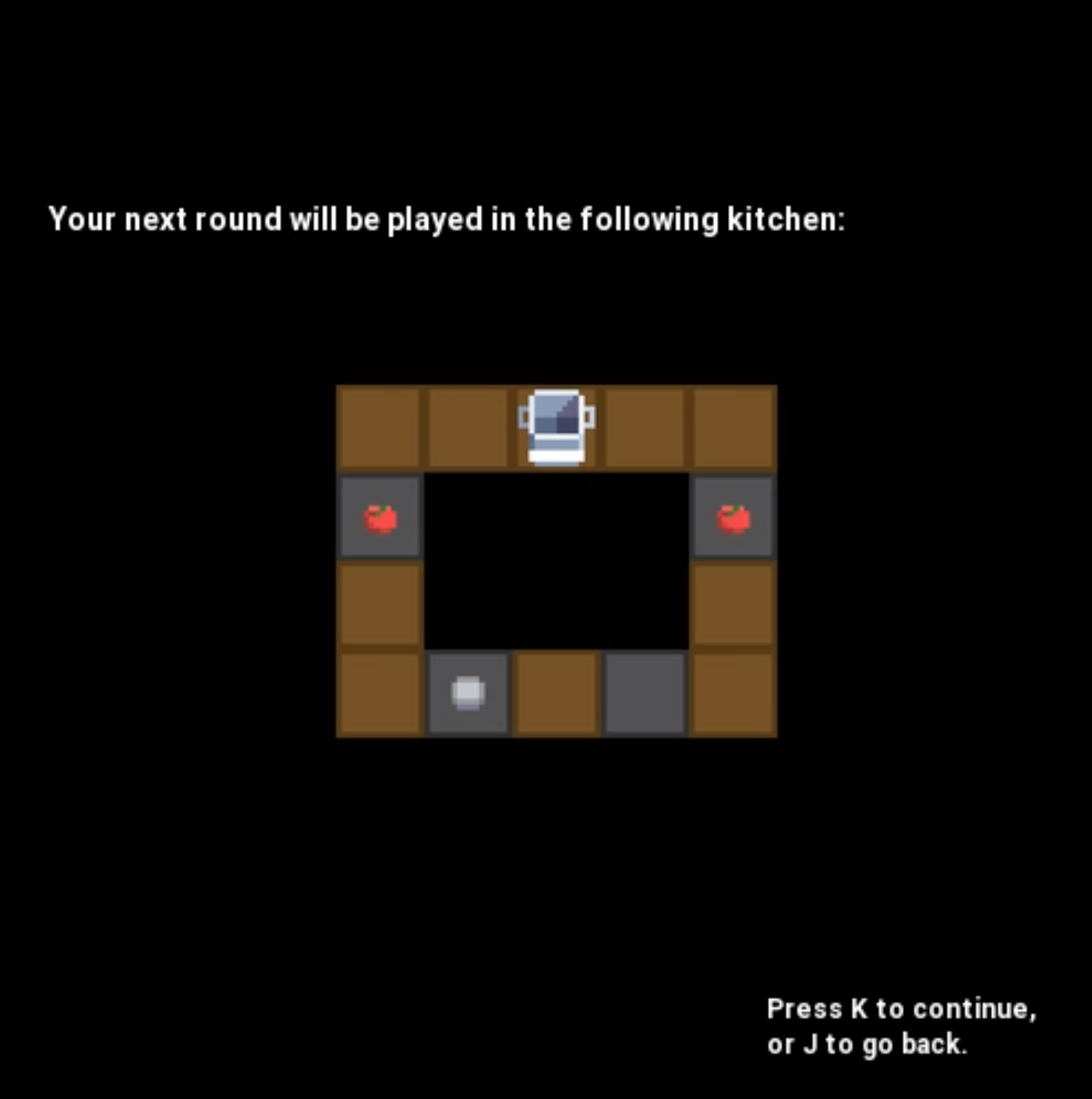}
        \caption{Screen 18: Introduce kitchen layout for first episode.}
    \end{subfigure}
    \hfill
    \begin{subfigure}{0.45\textwidth}
        \centering
        \includegraphics[width=\linewidth]{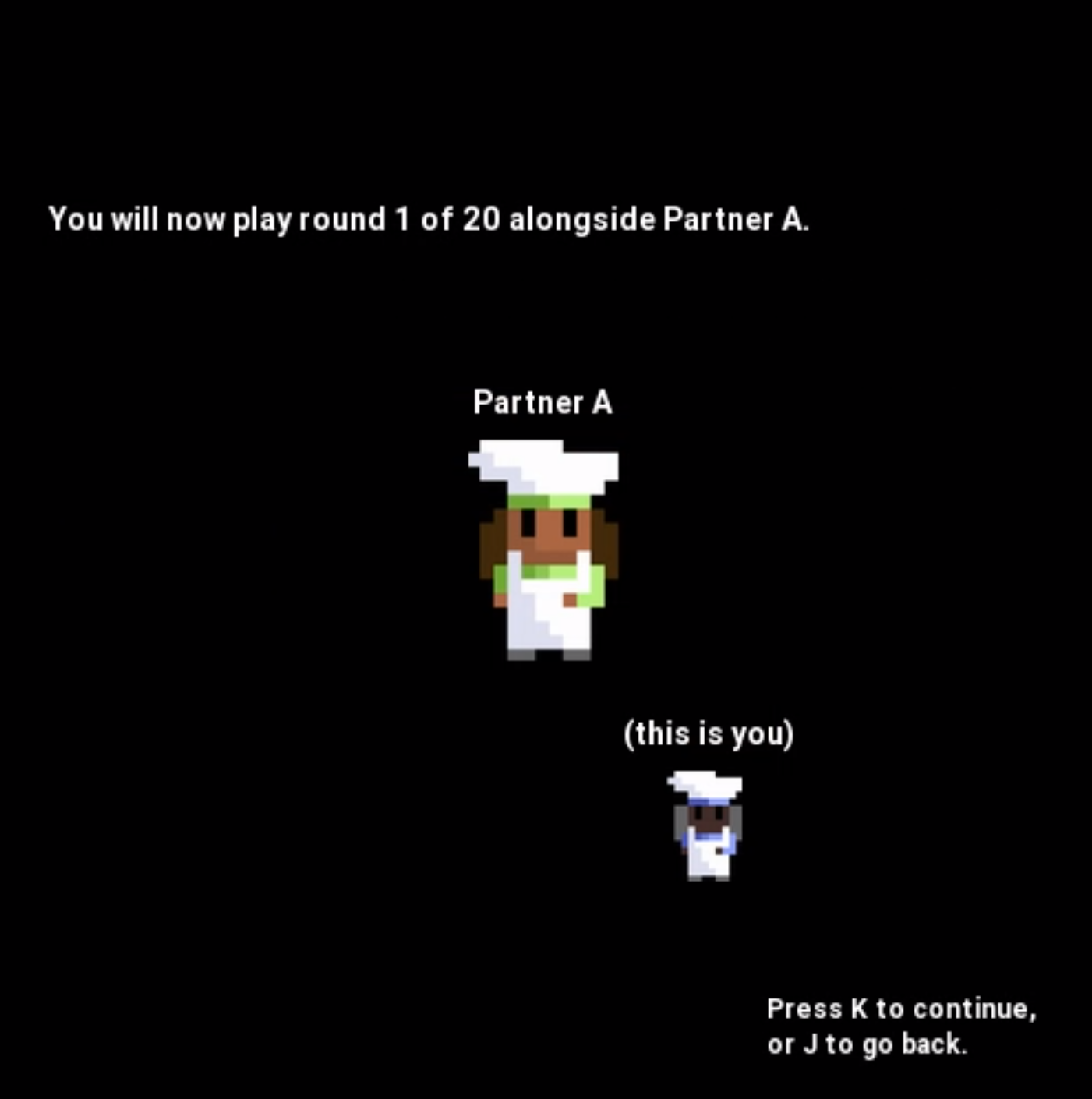}
        \caption{Screen 19: Introduce partner for first episode.}
    \end{subfigure}
    \hfill
    \begin{subfigure}{0.45\textwidth}
        \centering
        \includegraphics[width=\linewidth]{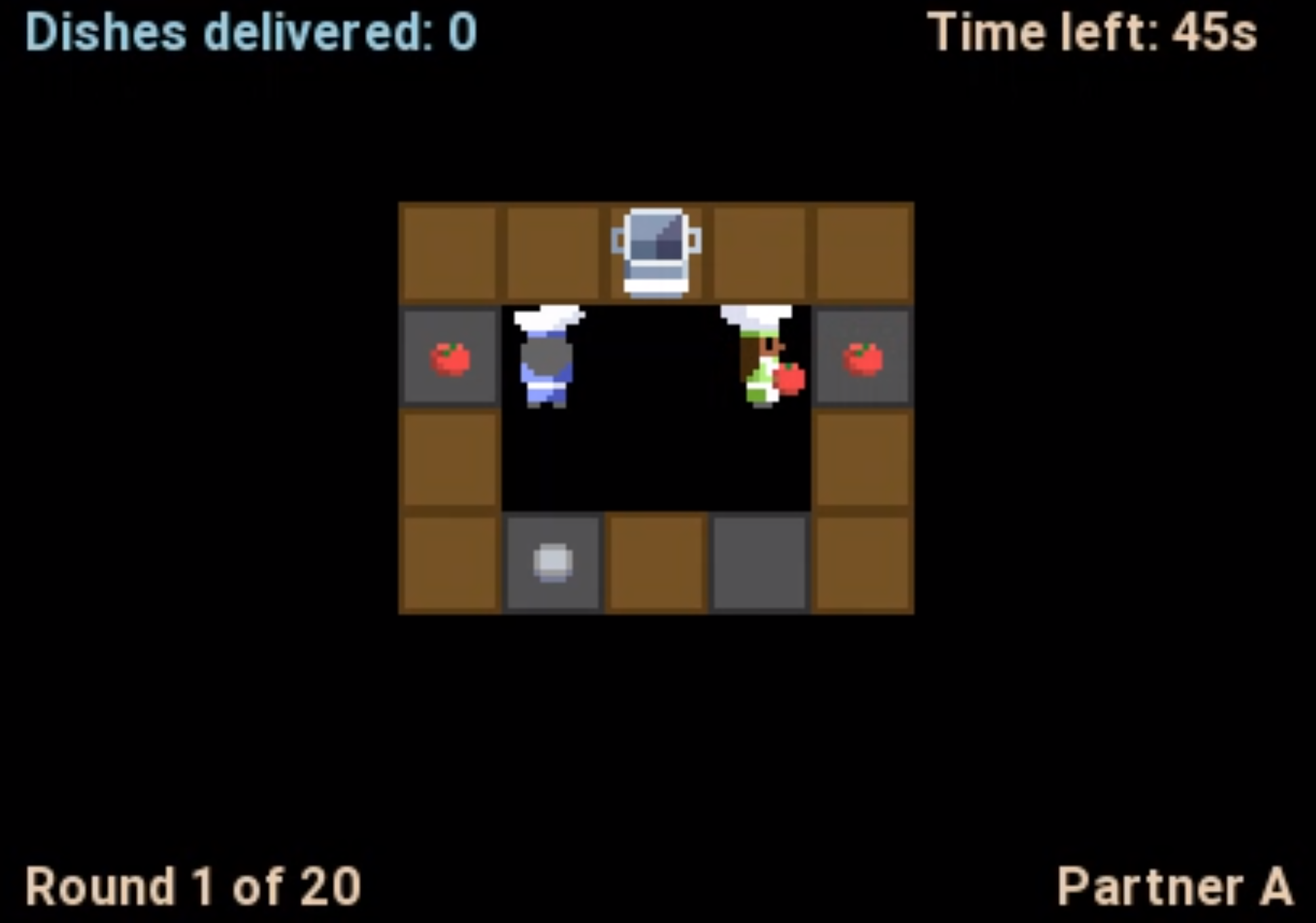}
        \caption{Screen 20: First episode.}
    \end{subfigure}
    \hfill
    \begin{subfigure}{0.45\textwidth}
        \centering
        \includegraphics[width=\linewidth]{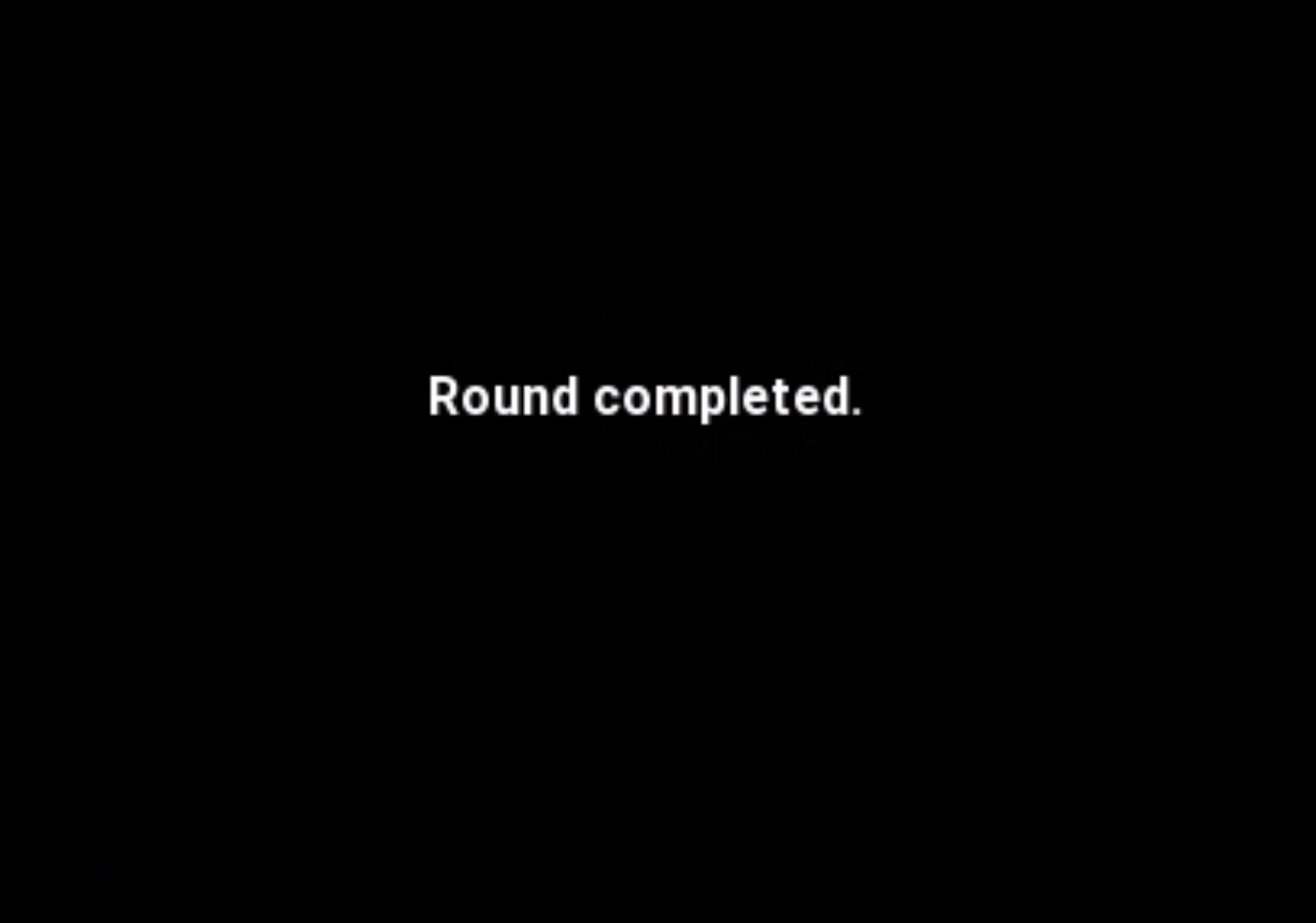}
        \caption{Screen 21: Confirm completion of first episode.}
    \end{subfigure}
    \caption{Screenshots of first episode.}
    \label{fig:app/screenshots_5}
\end{figure*}

\begin{figure*}
    \centering
    \begin{subfigure}{0.45\textwidth}
        \centering
        \includegraphics[width=\linewidth]{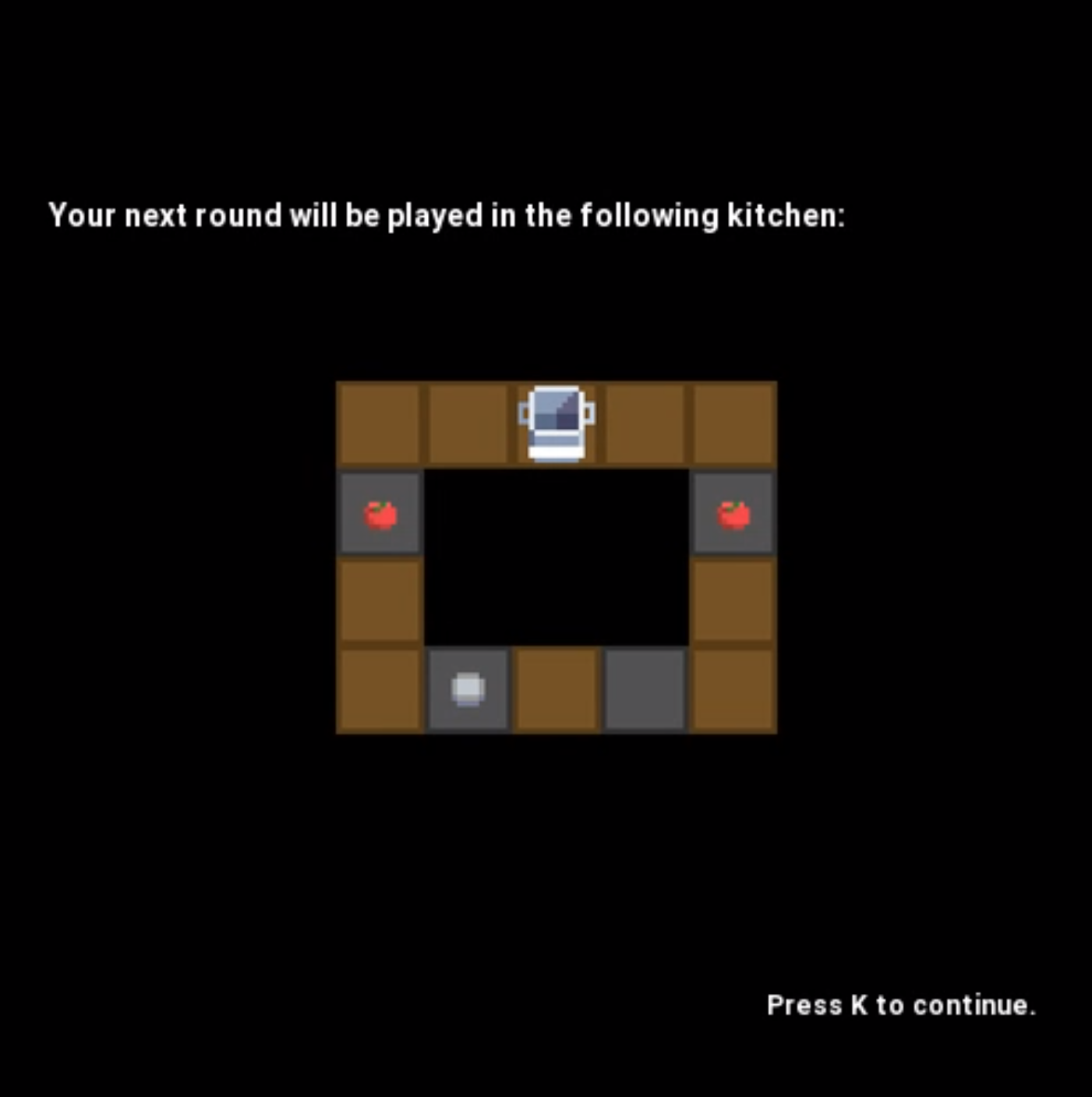}
        \caption{Screen 22: Introduce kitchen layout for second episode.}
    \end{subfigure}
    \hfill
    \begin{subfigure}{0.45\textwidth}
        \centering
        \includegraphics[width=\linewidth]{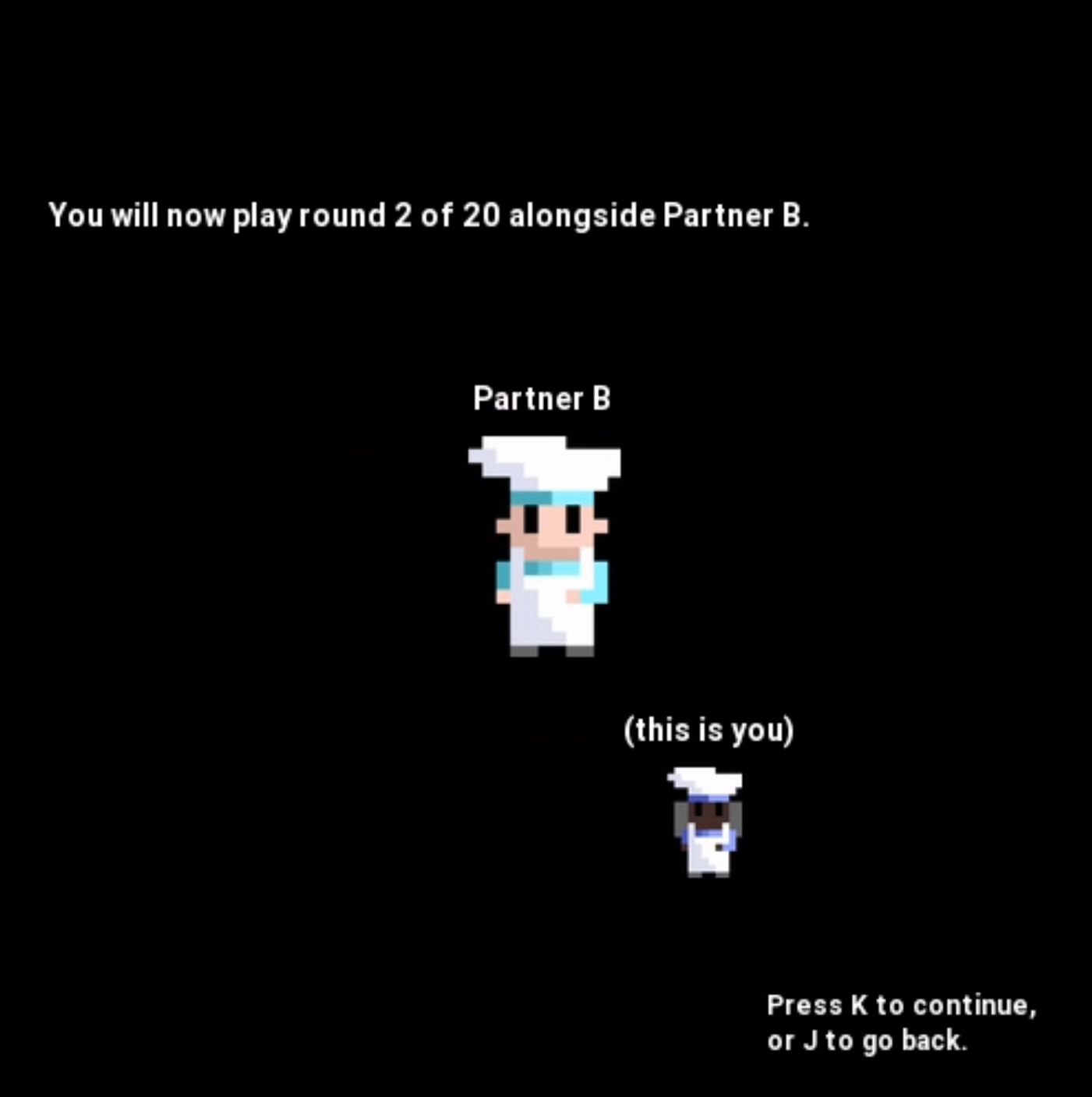}
        \caption{Screen 23: Introduce partner for second episode.}
    \end{subfigure}    
    \hfill
    \begin{subfigure}{0.45\textwidth}
        \centering
        \includegraphics[width=\linewidth]{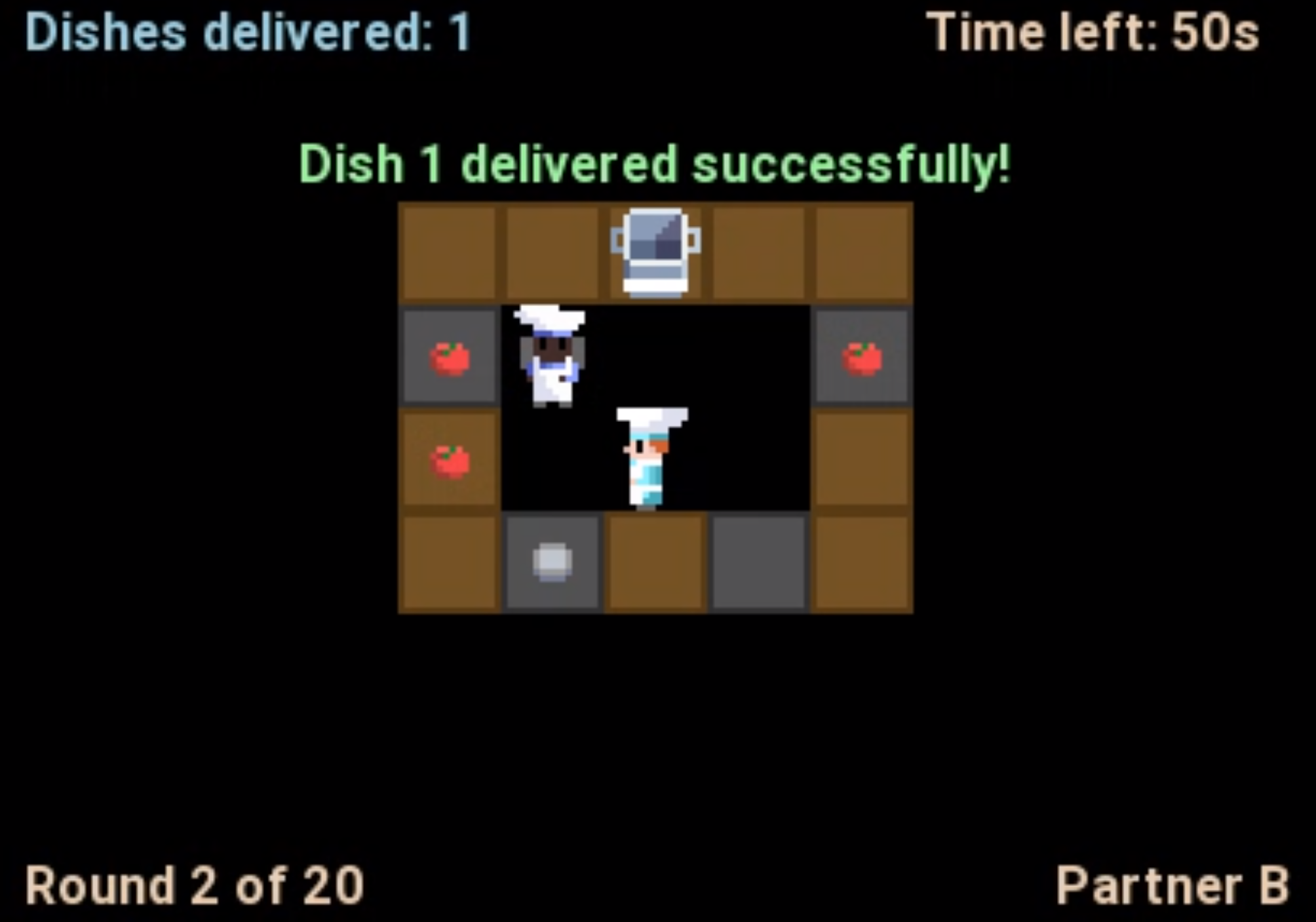}
        \caption{Screen 24: Second episode.}
    \end{subfigure}
    \hfill
    \begin{subfigure}{0.45\textwidth}
        \centering
        \includegraphics[width=\linewidth]{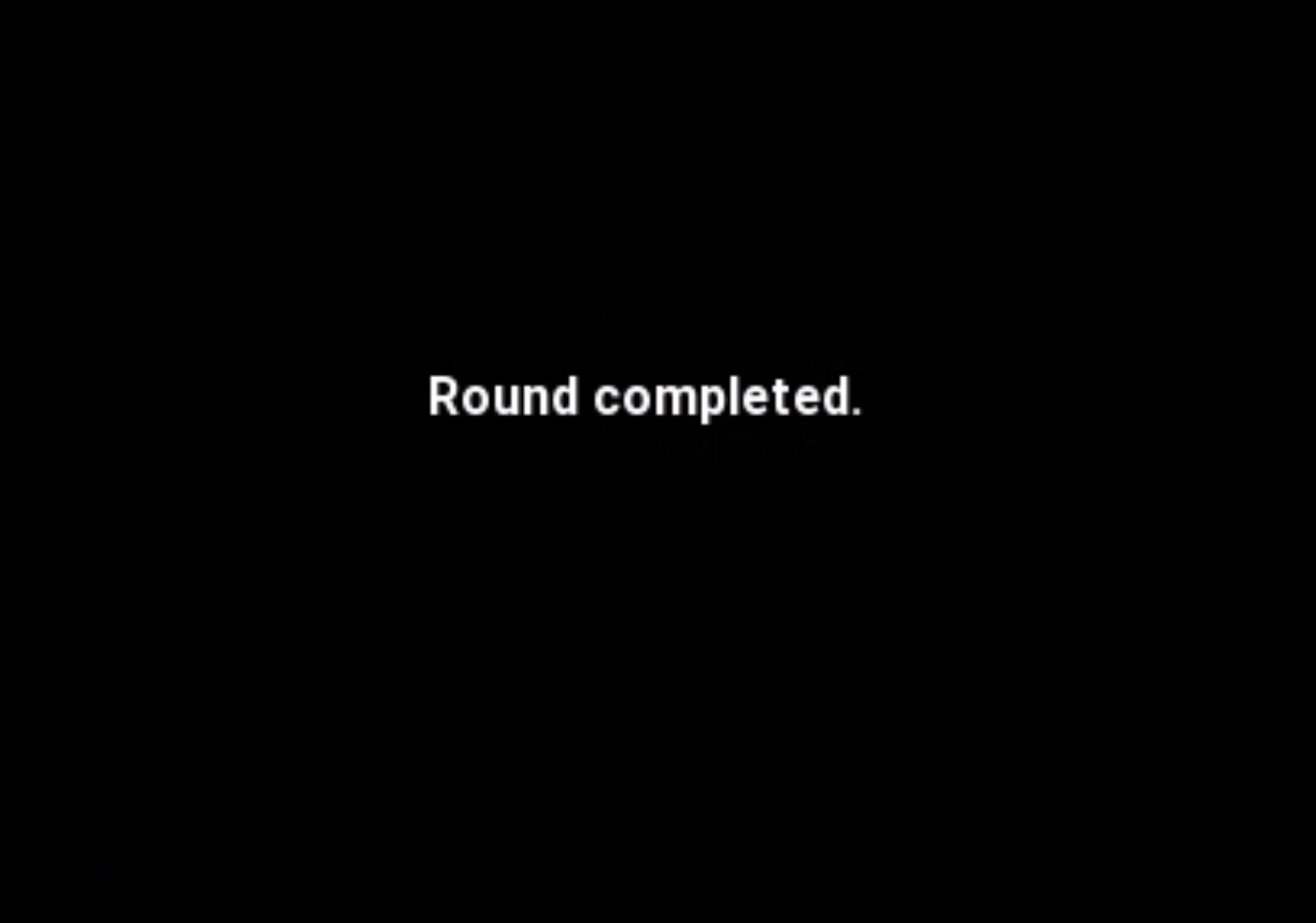}
        \caption{Screen 25: Confirm completion of second episode.}
    \end{subfigure}
    \caption{Screenshots of second episode.}
    \label{fig:app/screenshots_6}
\end{figure*}

\begin{figure*}
    \centering
    \begin{subfigure}{0.45\textwidth}
        \centering
        \includegraphics[width=\linewidth]{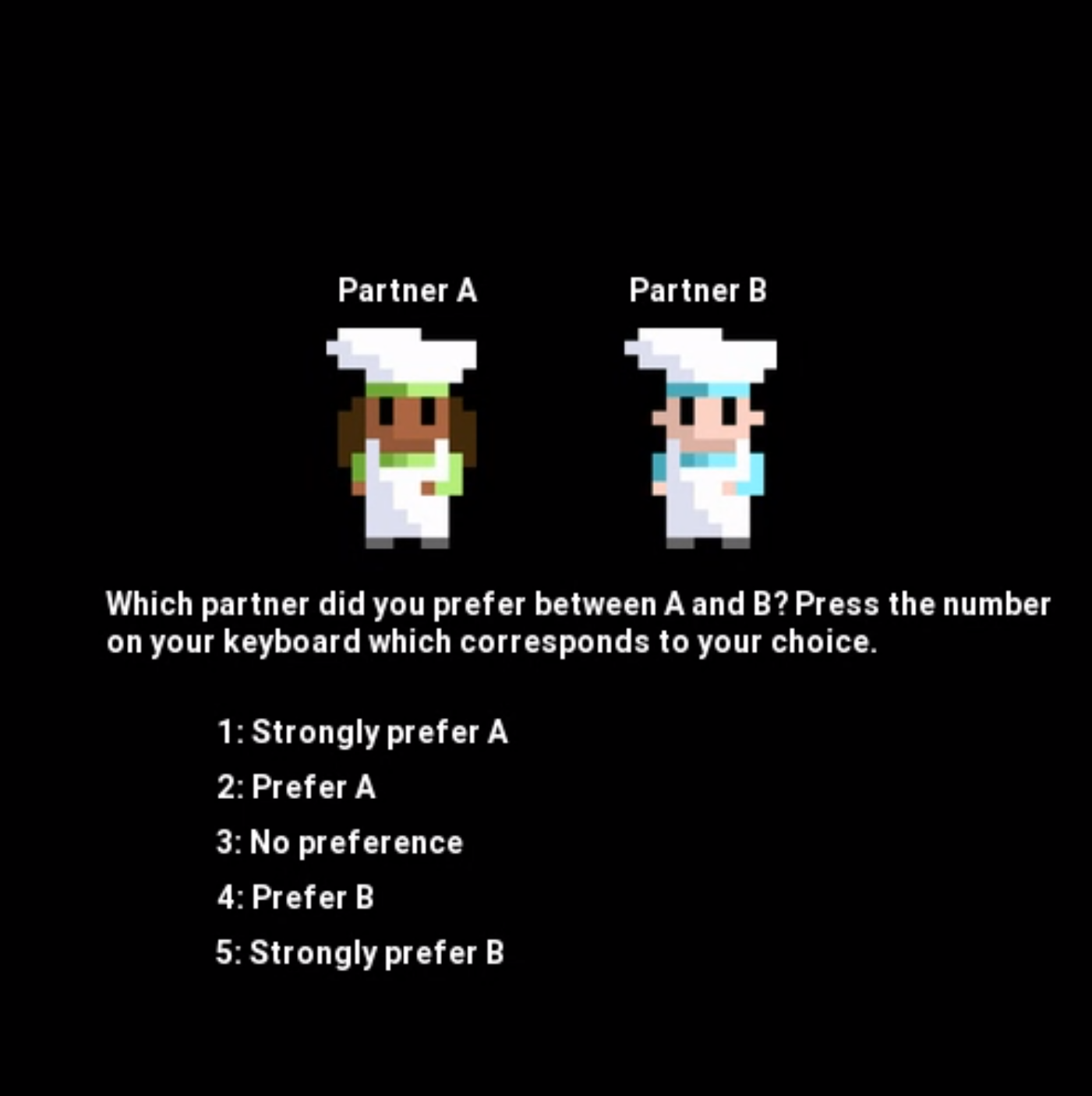}
        \caption{Screen 26: Elicit participant's preference \\ between partners.}
    \end{subfigure}    
    \hfill
    \begin{subfigure}{0.45\textwidth}
        \centering
        \includegraphics[width=\linewidth]{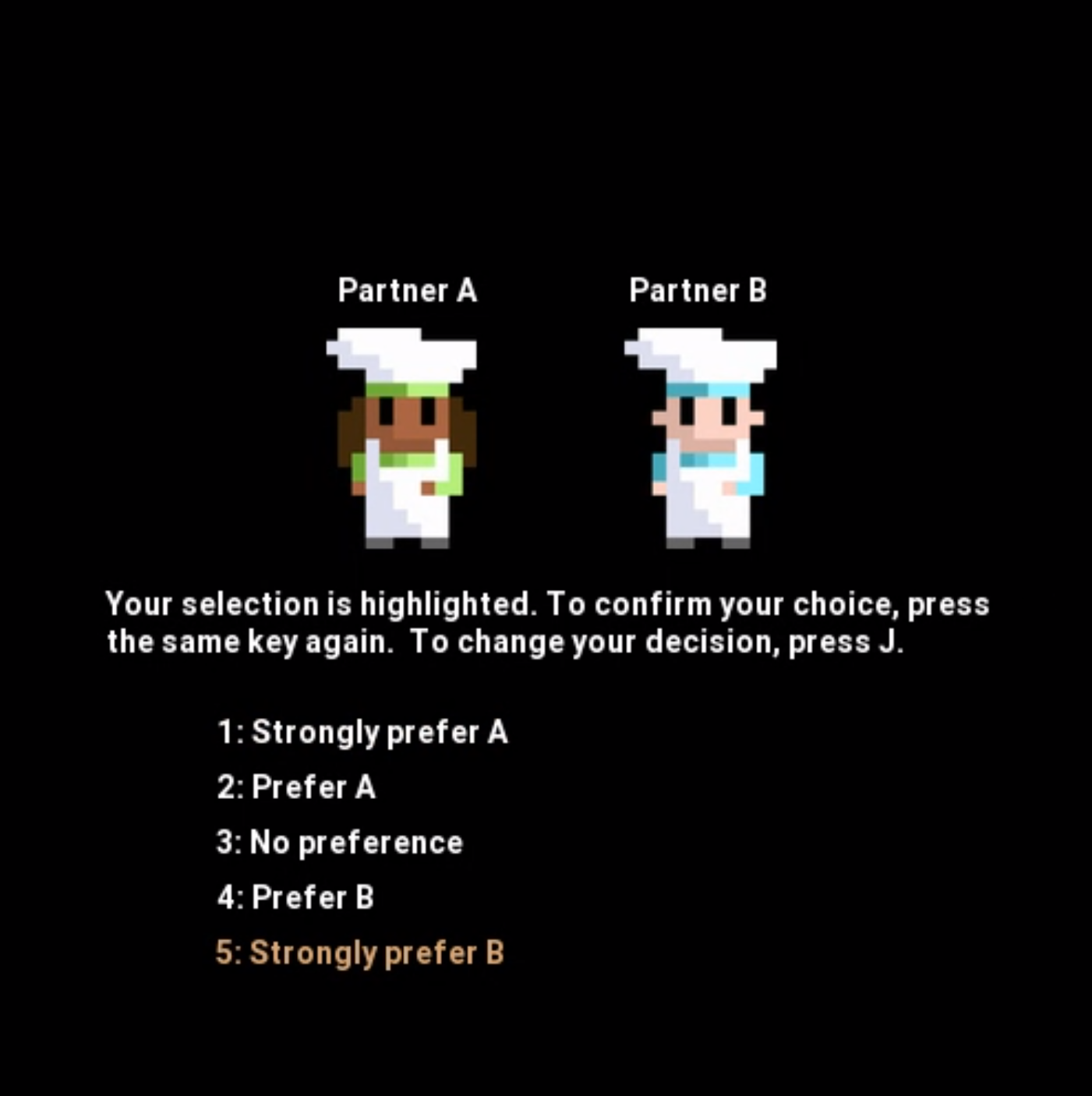}
        \caption{Screen 27: Confirm participant's preference between partners.}
    \end{subfigure}  
    \caption{Screenshots of preference elicitation over first and second episodes.}
    \label{fig:app/screenshots_7}
\end{figure*}

Here we include screenshots of our human-agent collaborative study:

\begin{enumerate}
    \item Instruction and tutorial screens (Figures~\ref{fig:app/screenshots_1} and \ref{fig:app/screenshots_2}).
    \item Playing a solo practice episode (Figure~\ref{fig:app/screenshots_3}).
    \item Instructions on the episodes with a partner (Figure~\ref{fig:app/screenshots_4}).
    \item Playing episode 1 with Partner A (Figure~\ref{fig:app/screenshots_5}).
    \item Playing episode 2 with Partner B (Figure~\ref{fig:app/screenshots_6}).
    \item Preference elicitation between Partners A and B (Figure~\ref{fig:app/screenshots_7}).
    \item Repeat steps 4-6 for 18 more episodes.
\end{enumerate}

\clearpage
\subsection{Analytic details}
\label{sec:app/human_ai/analytic_details}

We run a repeated-measures analysis of variance (ANOVA) to compare average team deliveries for each agent partner, with a random effect incorporated for each participant. Pairwise contrasts (using the Tukey method to adjust for multiple comparisons) indicated that the FCP agent achieved significantly higher delivery totals relative to the BCP agent, $t(1707) = 7.8$ ($p < 0.001$), the PP agent, $t(1707) = 8.2$ ($p < 0.001$), and the SP agent $t(1707) = 12.3$ ($p < 0.001$). The BCP agent also scored significantly higher than the SP agent, $t(1707) = 4.5$ ($p < 0.001$).

We conduct the same analysis to compare FCP against the FCP$_{-T}$ ablation. Human-agent teams involving the FCP agent completed significantly more deliveries than did those with the FCP$_{-T}$ agent, $t(797) = 14.4$ ($p < 0.001$).

To test whether participants preferred the FCP agent over other agent partners, we similarly fit a repeated-measures ANOVA on preferences elicited between FCP and any other agent, including the identity of the other agent as the sole predictor variable, as well as a random effect for each participant. Participants expressed significantly greater preferences for FCP over the SP agent, $t(329.6) = 4.9$ ($p < 0.001$), the PP agent, $t(399.0) = 2.5$ ($p = 0.011$), the BCP agent, $t(343.1) = 2.0$ ($p = 0.047$), and the FCP$_{-T}$ agent, $t(384.7) = 3.0$ ($p = 0.003$).

Finally, we run a repeated-measures ANOVA (again with a random effect for participants) to understand participants' preferences between the BCP agent and the PP agent. Participants reported significantly favoring the BCP agent over the PP agent, $t(77.8) = 3.1$ ($p = 0.003$).

\clearpage
\subsection{Additional quantitative results}
\label{sec:app/human_ai/quantitative_results}

\begin{figure}[h]
     \centering
     \begin{subfigure}[b]{\textwidth}
         \centering
         \includegraphics[width=\textwidth]{figures/environment/layouts.pdf}
         \caption{Layouts for reference.}
     \end{subfigure}
     \hfill
     \begin{subfigure}[b]{\textwidth}
         \centering
         \includegraphics[width=\textwidth]{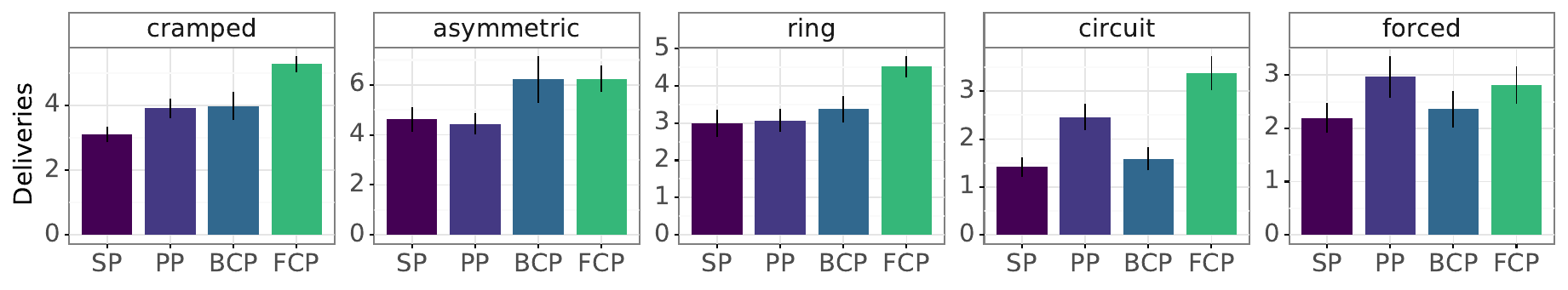}
         \caption{Human-agent team deliveries by map (FCP vs baselines).}
         \label{fig:deliveries_by_map}
     \end{subfigure}
     \hfill
     \begin{subfigure}[b]{\textwidth}
         \centering
         \includegraphics[width=\textwidth]{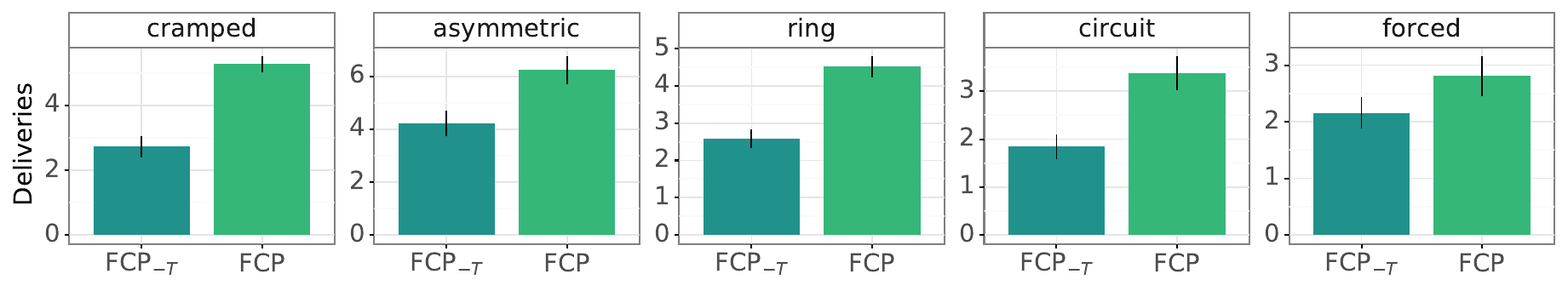}
         \caption{Human-agent team deliveries by map (FCP vs FCP$_{-T}$).}
         \label{fig:deliveries_by_map_fcp}
     \end{subfigure}
     \hfill
     \begin{subfigure}[b]{\textwidth}
         \centering
         \includegraphics[width=\textwidth]{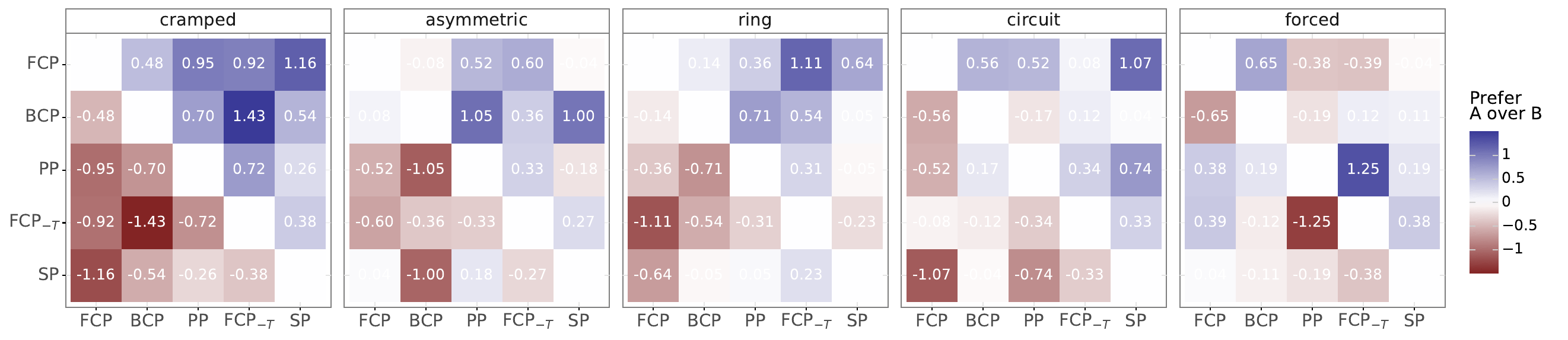}
         \caption{Human preferences for agent partners by map.}
         \label{fig:preferences_by_map}
     \end{subfigure}
     \hfill
    \caption{\textbf{Human-agent collaborative evaluation, per-layout:} Performance of each agent when partnered with human participants, as assessed by both objective and subjective metrics. In (a) and (b), error bars represent 95\% confidence intervals. In terms of deliveries, FCP performs comparably or better than all baselines on every map. FCP is also the most consistently preferred agent across maps, though on some specific maps (asymmetric and forced) participants occasionally report preferring other agents over FCP.}
\end{figure}

\subsection{Additional qualitative results}
\label{sec:app/human_ai/qualitative_results}

At the end of the study, we prompted participants with an open-ended question for feedback on their partners. Many responses touched on adaptability, specialization, and goal compatibility. We include a sample of responses below.

\begin{itemize}
    \item ``Some did not adapt well to what I was doing or they were chaotic and I could not find a way to adapt to them.''
    \item ``Sometimes, depending on the layout of the kitchen, my partner seemed to chose a role and stick with it, allowing me to take the other role, and this seemed to be the most efficient way to work. For example, if my partner was `in charge' of filling the pot, and I was in charge of putting it on a plate and delivering it. Also, I noticed that some of my partners seemed to know they needed to move around me, while others seemed to get `stuck' until I moved out of their way.''
    \item ``I prefer partners which were willing not just to work independently but to let me hand them tomatoes/plates (since I maneuvered slower than them). It was difficult when they did not work on the same goal together but acted independently.''
    \item ``The responsive and compatibility of the other partners was really interesting as it took some quite some time to figure out what strategy works best for us and the strategy was almost instant with other partners.''
    \item ``Oh man I never laughed so much in my life. My partners were all fun to play with. Some were a bit faster than others. I think the one in pink was the fastest overall.''
\end{itemize}

\section{Related work}
\label{sec:app/related_work}
Our work is similar to that conducted by \citet{carroll2019overcooked}. Here we provide a summary of the notable differences in experimental design that may contribute to differences in our results.

First, we did not use population-based training (PBT) with $N=3$ as a baseline. We instead used population-play (PP) with $N=32$, which shares some similarities with PBT. In our experiments, PP performed significantly better than SP. One potential reason for the performance difference between these two experiments is the small $N=3$ population size used for PBT; this may have led to a collapse of diversity. This echos a key finding from Jaderberg et al.'s \citepa{jaderberg2017pbt} methodological evaluation of PBT: ``If the population size is too small (10 or below) we tend to encounter higher variance and can suffer from poorer results``. This issue may have been exacerbated by the two-player common-payoff nature of our task.

Second, we trained our agents on an egocentric observation of the environment, as opposed to a top-down layer representation of the whole environment. Egocentric observations can improve agent generalization \citepa{ye2020rotation}. The fixed observation size in our approach additionally allowed us to train a single agent to play on all layouts, in contrast with Carroll et al.'s approach of training one agent per layout. We also used the VMPO learning algorithm \citep{song2019vmpo} rather than PPO.

We implemented the game environment in the open-source environment engine DMLab2D \citepa{beattie2020dmlab2d}.
Our implementation followed the implementation reported in the \citet{carroll2019overcooked}, but was likely not a perfect recreation. We implemented our own human-agent interaction pipeline on top of the DMLab2D environment, with additional instructions and text to explain the game (as described in Section~\ref{sec:app/human_ai}). In our human-agent experiments, episodes lasted 300 steps over 60 seconds ($5$ FPS); in \citet{carroll2019overcooked}, episodes lasted 400 steps over 60 seconds ($6.66$ FPS).

Finally, our BC agents were trained on 12,000 environment steps per layout, rather than 18,000 steps per layout, due to data collection limitations. We used an architecture with larger layers and an overall higher number of layers than used in \citet{carroll2019overcooked}. The observable features used to train the BC agent differed slightly; nonetheless, we observed no noticeable difference in performance. With our BC approach, we found that the random-action heuristic \citet{carroll2019overcooked} used was unnecessary.

\bibliographya{main}

\end{document}